\documentclass[sigconf, 9pt]{acmart}
\usepackage{bbding}
\usepackage{amsmath}
\usepackage{color}
\usepackage{dsfont}
\usepackage{multirow}
\usepackage{makecell}
\usepackage{caption}
\usepackage{subcaption}
\usepackage{enumitem}
\usepackage[symbol]{footmisc}

\copyrightyear{2023} 
\acmYear{2023} 
\setcopyright{acmlicensed}\acmConference[SenSys '23]{The 21st ACM Conference on Embedded Networked Sensor Systems}{November 12--17, 2023}{Istanbul, Turkiye}
\acmBooktitle{The 21st ACM Conference on Embedded Networked Sensor Systems (SenSys '23), November 12--17, 2023, Istanbul, Turkiye}
\acmPrice{15.00}
\acmDOI{10.1145/3625687.3625793}
\acmISBN{979-8-4007-0414-7/23/11}

\begin{document}

\title{EdgeFM: Leveraging Foundation Model for Open-set Learning on the Edge}

\author{Bufang Yang}
\affiliation{
  \institution{
    The Chinese University of Hong Kong \city{Hong Kong SAR}\country{China\linebreak bfyang@link.cuhk.edu.hk}}
}

\author{Lixing He}
\affiliation{
  \institution{
    The Chinese University of Hong Kong \city{Hong Kong SAR}\country{China\linebreak 1155170464@link.cuhk.edu.hk}}
}

\author{Neiwen Ling}
\affiliation{
  \institution{
    The Chinese University of Hong Kong \city{Hong Kong SAR}\country{China\linebreak lingnw@link.cuhk.edu.hk}}
}

\author{Zhenyu Yan$^{\dag}$}\thanks{$^{\dag}$ Corresponding author}
\affiliation{
  \institution{
    The Chinese University of Hong Kong \city{Hong Kong SAR}\country{China\linebreak zyyan@cuhk.edu.hk}}
}


\author{Guoliang Xing}
\affiliation{
  \institution{
    The Chinese University of Hong Kong \city{Hong Kong SAR}\country{China\linebreak glxing@cuhk.edu.hk}}
}

\author{Xian Shuai}
\affiliation{
  \institution{
    Noah’s Ark Lab, Huawei Technologies \city{Hong Kong SAR}\country{China\linebreak shuaixian1@huawei.com}}
}

\author{Xiaozhe Ren}
\affiliation{
  \institution{
    Noah’s Ark Lab, Huawei Technologies \city{Hong Kong SAR}\country{China\linebreak renxiaozhe@huawei.com}}
}

\author{Xin Jiang}
\affiliation{
  \institution{
    Noah’s Ark Lab, Huawei Technologies \city{Hong Kong SAR}\country{China\linebreak Jiang.Xin@huawei.com}}
}

\renewcommand{\shortauthors}{B. Yang, L. He, N. Ling, Z. Yan, G. Xing, X. Shuai, X. Ren, and X. Jiang}


\begin{abstract}
Deep Learning (DL) models have been widely deployed on IoT devices with the help of advancements in DL algorithms and chips.
However, the limited resources of edge devices make these on-device DL models hard to be generalizable to diverse environments and tasks.
Although the recently emerged foundation models (FMs) show impressive generalization power, how to effectively leverage the rich knowledge of FMs on resource-limited edge devices is still not explored.
In this paper, we propose EdgeFM, a novel edge-cloud cooperative system with open-set recognition capability.
EdgeFM selectively uploads unlabeled data to query the FM on the cloud and customizes the specific knowledge and architectures for edge models.
Meanwhile, EdgeFM conducts dynamic model switching at run-time taking into account both data uncertainty and dynamic network variations, which ensures the accuracy always close to the original FM.
We implement EdgeFM using two FMs on two edge platforms. We evaluate EdgeFM on three public datasets and two self-collected datasets.
Results show that EdgeFM can reduce the end-to-end latency up to 3.2x and achieve 34.3\% accuracy increase compared with the baseline.

\end{abstract}



\settopmatter{printfolios=true}
\maketitle

\section{Introduction}





Deep learning models have been widely deployed on IoT systems thanks to their excellent performance and the advancement in edge Artificial intelligence (AI) hardware. There will be more than 1.5 billion edge AI processors shipped in 2024 \cite{EdgeAIchip}. Various commercial and industrial applications are deployed on embedded AI systems, such as health monitoring systems \cite{ouyang2022cosmo,yang2022novel}, service and logistics robots \cite{shah2022lm}, and sound event detection systems \cite{xie2021zero,islam2019soundsemantics}. As these applications must operate in complicated and ever-changing environments, scalability and adaptiveness are of great importance.

Most current on-device AI models are task-specific and can only predict a \textit{closed-set} of classes pre-defined at the training stage \cite{ouyang2022cosmo,xu2021limu,yang2021single}.
Their performance degrades severely when the class of input is not seen during the training.
Although various approaches such as transfer learning \cite{akbari2019transferring,mathur2019unsupervised} and meta-learning \cite{gong2019metasense, ding2020rf} have been proposed to calibrate models on the edge, they still require non-trivial efforts for manual data annotation and on-device training, which are not practical in real-world deployments
\cite{meyer2019importance,ouyang2022cosmo,kishida2021object}.
The recent emergence of foundation models (FMs) 
such as GPT \cite{brown2020language} and CLIP \cite{radford2021learning} have shown impressive general knowledge that can support diverse downstream tasks like image captioning, question answering, and information extraction.
Current commercial proprietary FMs usually contain millions or billions of parameters and are pre-trained using billions of data samples \cite{bommasani2021opportunities,li2023freelm}.
Moreover, multi-modal FMs such as CLIP and ImageBind can learn embedded matching through massive paired data of different modalities, such as text, images, audio, and motion data. Their general knowledge can process diverse sensor data with \textbf{open-set recognition} capability when the interested types of classes change.



\begin{figure}[tbp]
  \centering
\includegraphics[width=1\linewidth]{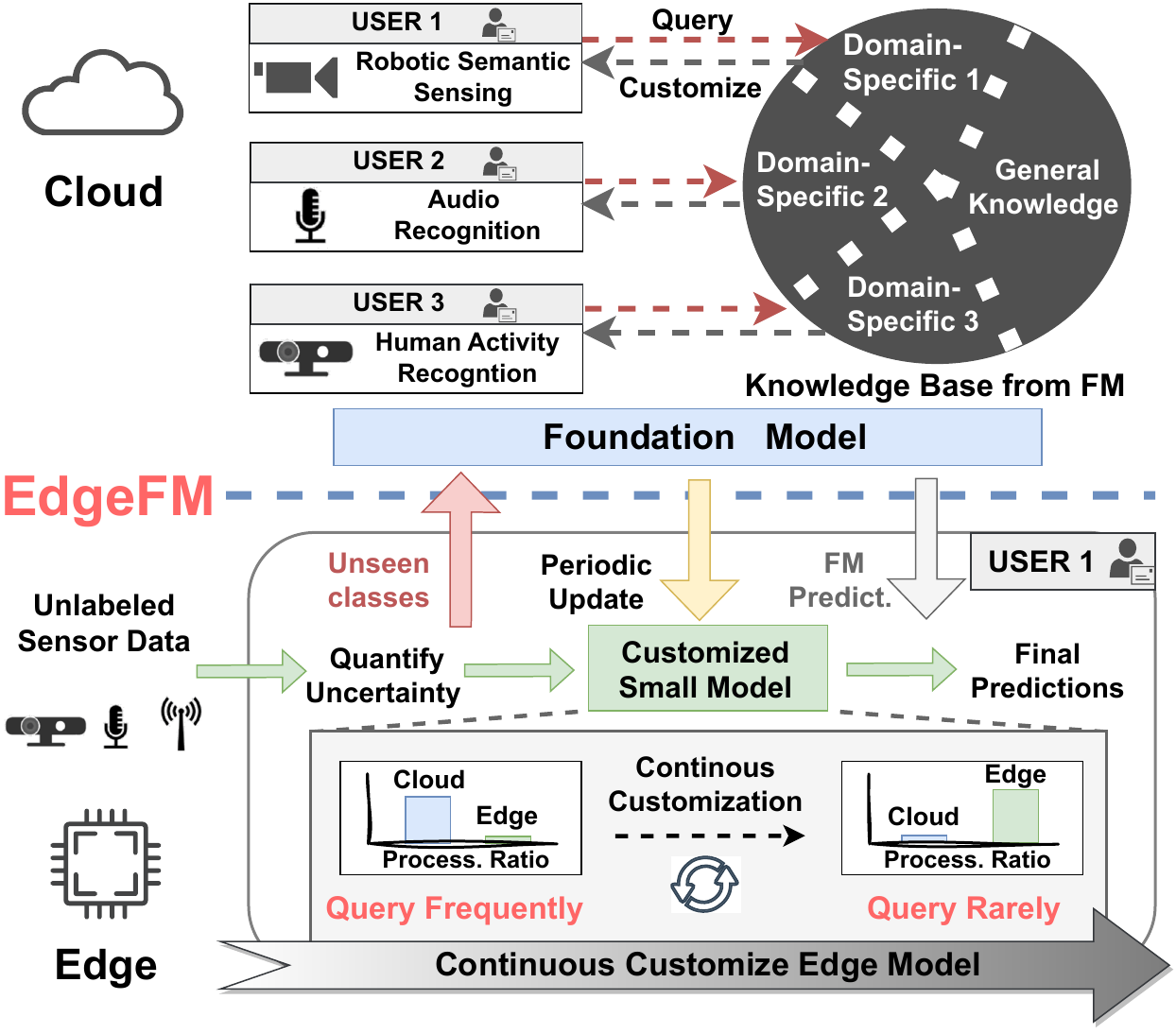}
  \caption{
An example of EdgeFM, enabling edge devices with open-set capability using dynamic customization and runtime model switching.
}
  \label{fig:overview1}
    \vspace{-1.5em}
\end{figure}

Some systems execute FMs on the cloud for tasks like robotic navigation \cite{liang2022effective,shah2022lm,ahn2022can}. However, transmitting all raw data to the cloud is not feasible in many practical scenarios.
Executing bulky FMs on the edge directly is infeasible due to limited resources. Current model compression techniques \cite{sun2023dime,wei2022contrastive,mobile_sam} treat all samples equally during inference, which leads to significant performance degradation for difficult/unseen input data.
Several studies partition a large model and deploy them to the edge and cloud for collaborative execution \cite{kang2017neurosurgeon,laskaridis2020spinn,zhao2021edgeml}. Most FMs adopt transformers whose output is even larger than the data input \cite{vaswani2017attention}, so model partitioning of FMs is not desirable.
In this paper, we propose EdgeFM, a novel edge-cloud cooperative system that can achieve open-set recognition capability by leveraging FMs for selective knowledge query and edge model customization.
As shown in Figure~\ref{fig:overview1}, FM is deployed on a cloud server and acts as a knowledge base containing both general and domain-specific knowledge.
At the inference stage, EdgeFM first determines whether it should query FMs based on the uncertainty of semantic features of sensor data and the real-time network variations, which ensures the accuracy is always close to the original FM.
Meanwhile, EdgeFM selectively uploads unlabeled data to query the FM on the cloud and periodically customizes the domain-specific knowledge and architectures for small models in a \textit{label-free manner}.
When the data distribution or interested class set changes, EdgeFM will query the knowledge from FMs frequently at the early stage, while it can primarily execute customized small models on edge devices at the late stage, thus reducing system overhead subsequently.
EdgeFM supports different tasks and modalities so that different users can query their domain-specific knowledge from FM while only executing their customized small models on resource-constrained edge devices to save system overhead.

We extensively evaluate the performance of EdgeFM on two FMs (CLIP and ImageBind), and two edge devices (Nvidia Jetson Xavier and Nano), using three public datasets and two self-collected datasets.
Our results show that EdgeFM reduces end-to-end latency up to 3.2x compared with existing on-device inference approaches.
EdgeFM can also achieve up to 34.3\% accuracy increase compared with the existing on-device open-set recognition approach.

The contributions of this paper are summarized as follows:
\begin{itemize}
  \item We propose EdgeFM, the first edge-cloud cooperative system with open-set recognition capability leveraging FMs for selective knowledge query and dynamic edge model customization. The system can work with sensor data of different modalities.
  


  \item We design a semantic-driven customization approach that allows EdgeFM to customize the domain-specific knowledge and the architectures of mobile-friendly models in a \textit{label-free manner}.

  \item We develop a dynamic model switching approach considering both the uncertainty of semantic features of sensor data and the real-time network fluctuation.

  \item 
  We implement EdgeFM on two FMs and deploy it to PC and two edge platforms. The evaluation includes three public datasets and two real-world datasets collected by ourselves about daily activity recognition and robot semantic recognition. EdgeFM can reduce the end-to-end latency up to 3.2x and achieve a 34.3\% accuracy increase compared with the baseline.
  
\end{itemize}

\section{Background and RELATED WORK}
\subsection{Multi-Modal Foundation Models}
\textit{Foundation models (FMs)} refer to a new class of large machine learning models that can extract valuable features to support diverse downstream tasks such as chatbot (e.g., GPT \cite{brown2020language}) and image recognition (e.g., CLIP \cite{radford2021learning}). Multi-modal FMs represented by CLIP and ImageBind \cite{radford2021learning,girdhar2023imagebind} learn the pairing of data of different modalities (e.g., RGB image, depth image, and audio) and their corresponding textual description across the internet to achieve the capability of \textit{open-set recognition}.
In particular, multi-modal FMs adopt a transformer-based encoder to extract features from the raw data and convert them into \textit{embedding}.
FMs also use a text encoder to extract \textit{text embedding} from the corresponding textual description of the raw data (e.g., images).
The training of such multi-modal FMs often adopts contrastive learning to study the pairing between the data embedding and the corresponding text embedding to construct a \textit{unified embedding space}. 
This is also the reason that multi-modal FMs like ImageBind can work with one or multi-modal sensor data.
Specifically, for any class described in a natural language manner, FMs such as CLIP first convert the class name $CLS$ into a text description through concatenating $CLS$ with a pre-defined text template called prompt, such as ``This is a photo of a \{CLS\}''.
Then, the text description and the sensor data (such as images) are fed into the text encoder and image encoder of FMs to obtain respective embedding.
After computing the similarity score between the text embedding and other sensor data's embedding, FMs select the class with the highest score as the final prediction.

\subsection{Related Works}
\textbf{Open-set Recognition.}
Open-set recognition aims to recognize any classes described in a natural language manner without fine-tuning.
Existing approaches for open-set recognition can be classified into two categories: semantic-based and generative adversarial network (GAN)-based.
Semantic-based approaches \cite{xie2021zero,su2022distinguishing,chen2021elaborative,islam2019soundsemantics,tong2021zero} build the connection between the embeddings of sensor data and the semantic embedding of natural language.
GAN-based approaches \cite{narayan2020latent} use a GAN-based semantic decoder to synthesize features for unseen classes and mix them up with the real features of seen classes for model training.
However, their training data and model parameters are still limited compared to FMs, showing unsatisfactory performance of open-set recognition on embedded edge systems.

\noindent\textbf{Efficient On-device Inference.}
Many studies are proposed to accelerate NN inference on edge devices, such as model compression \cite{wu2016quantized,PatDNN}, early-exit \cite{leontiadis2021s,MobiVQA}, input filtering \cite{InFi,mmSampler} and reusing \cite{drolia2017cachier,xu2018deepcache}.
However, model compression, such as quantization \cite{wu2016quantized}, pruning \cite{PatDNN}, and knowledge distillation (KD) \cite{hinton2015distilling}, 
are static acceleration approaches.
They will also suffer severe performance degradation when compressing FMs into the scale of lightweight CNNs.
Early exit \cite{leontiadis2021s,MobiVQA} can dynamically reduce the redundant computation of NNs.
However, the high-dimensional embeddings and deep layers of FMs make the early-exit heads very heavyweight.
Moreover, they still require executing the entire FM on the edge for hard samples, which can exceed the memory bound of edge devices like Nvidia Jetson Nano.
Some work proposes to optimize computation efficiency by processing input data, including filtering \cite{mmSampler,InFi} and reusing \cite{drolia2017cachier,xu2018deepcache}.
However, they can still not address the challenge of insufficient memory on edge devices due to the bulky size of FMs.


\noindent\textbf{FMs on the Edge.}
Although FMs are more recently emerging, some approaches have been proposed to optimize the inference of FMs.
MLC-LLM \cite{mlc-llm} leverages memory planning and quantization techniques to run LLMs on the phone.
Tabi \cite{wang2023tabi} and FrugalGPT \cite{chen2023frugalgpt} propose to cascade different sizes of models for acceleration.
However, these approaches are tailored for generating text dialogues, which cannot work with multi-modal sensor data.
A line of work including DIME-FM \cite{sun2023dime}, FD-CLIP \cite{wei2022contrastive}, VLKD \cite{dai2022enabling}, and MobileSAM \cite{mobile_sam} aims to compress multi-modal FMs by KD.
However, most of the previous works focus on preserving the great open-set ability of FMs rather than generating task-specific small models, thus always requiring heavyweight transformer-based architectures \cite{vaswani2017attention}, which is hard to be implemented on embedded systems.
These approaches also require the dedicated design of small model architectures, limiting their practicality.

\noindent\textbf{Edge-cloud Collaboration.}
Several works propose to adopt edge and cloud collaboration solutions to achieve the trade-off between accuracy and efficiency, 
Neurosurgeon \cite{kang2017neurosurgeon} splits the NN to deploy the several layers on the edge while the remaining layers are on the cloud.
SPINN \cite{laskaridis2020spinn} integrates the model splitting and early-exit to co-optimize multiple objectives, including accuracy and latency,
by adjusting the split point and early exit threshold.
AgileNN \cite{huang2022real} and DeepCOD \cite{yao2020deep} compress the size of transmitted intermediate features to improve the edge-cloud inference efficiency.
There are also edge-cloud cooperative systems based on big-little model-switching \cite{park2015big,li2021appealnet}.
However, most of the previous edge-cloud cooperative systems focus on
task-specific NNs, which work in a closed-set manner. The development of an open-set supported cooperative system remains unexplored.

In summary, existing works either focus on optimizing the on-device efficient inference for closed-set NNs, or compressing FMs through dedicated designs that are static and not scalable for dynamic real-world IoT applications.
Leveraging open-set knowledge of FMs for embedded systems is still an ill-address problem.

\section{Motivation: A Case Study}
The capability of open-set recognition is highly desirable in both commercial and industrial embedded systems. 
Current FM services predominantly adopt cloud-centric solutions \cite{liberty2020elastic,google}.
We first evaluate the performance of cloud-centric FM services under real-world dynamic network conditions.
Next, we measure the performance of FMs and small-size recognition models on objects of unseen classes. 
This applies to many embedded systems installing an AI model that can only recognize a limited set of objects due to constrained resources. We use a public image dataset FLO102 \cite{nilsback2008automated} with 102 types of flowers, and an image dataset collected by us containing 40 classes of common objects in the indoor environments. We use CLIP \cite{radford2021learning} and ImageBind \cite{girdhar2023imagebind} as FMs and use lightweight models for embedded systems like MobileNetV2 \cite{howard2017mobilenets} and ResNet18 \cite{he2016deep} as small models.
In particular, we first study the performance of FM and small models on unseen classes, the customization of small models to adapt to the dynamic set of classes, and the execution efficiency of FMs and small models. Then, we test the feasibility of customizing embedded models with FM's knowledge for the open-set capability.

\subsection{Cloud-centric FM services}
\begin{figure} 
    \centering 
    \includegraphics[width=0.48\textwidth]{{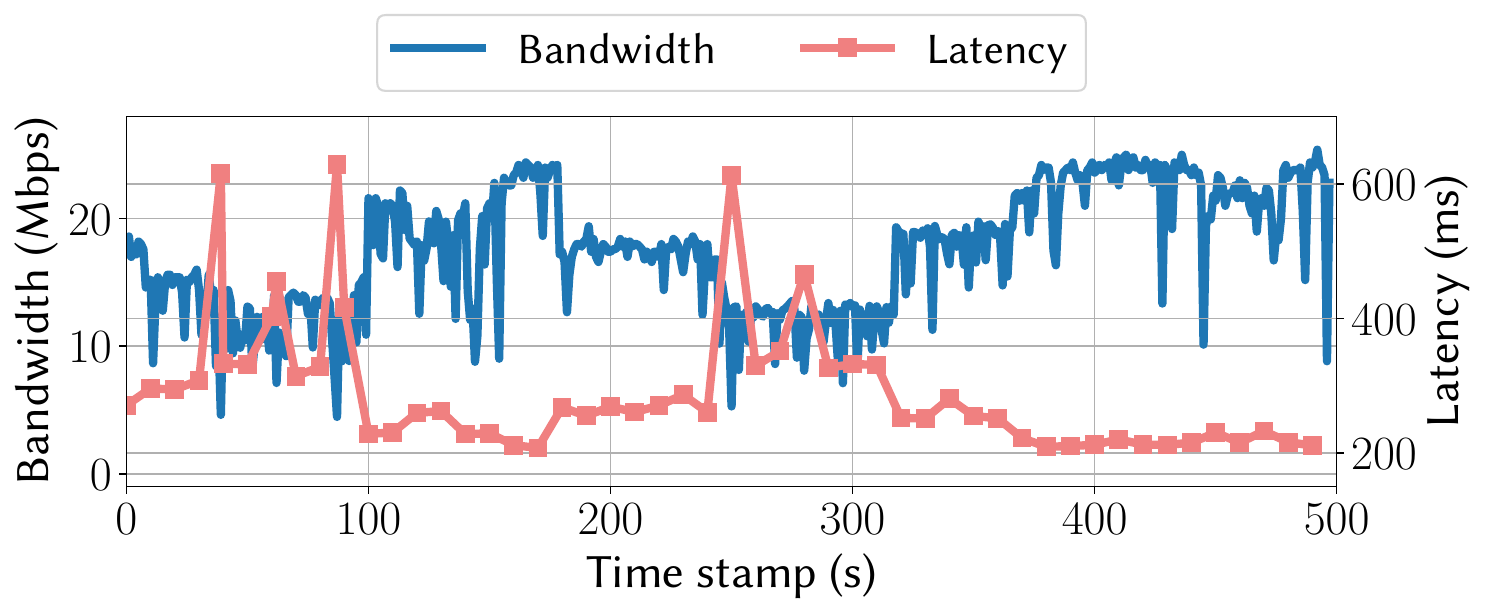}}
  \caption{
  An example showing the inference latency of cloud-centric FM solutions under dynamic network conditions.}
  \label{fig:e2e_cloudcentric}
   \vspace{-1.0em}
\end{figure}

We first measure how the cloud-centric FM service performs on an edge system. 
We set up an edge platform (i.e., NVIDIA Jetson Nano) to stream RGB images of an office room to the cloud server for recognizing the objects. The cloud server deploys ImageBind as the FM.
Figure~\ref{fig:e2e_cloudcentric} shows the network bandwidth measurements and corresponding inference latency. 
The inference latency exhibits considerable fluctuations under dynamic network conditions, ranging from 200 to 630 ms. 
However, cloud-centric approaches necessitate streaming all data to the cloud server, causing non-trivial system latency (up to 630 ms) due to varying network conditions. 
The increased delays can significantly affect user experience, such as the risk of collisions of home service robots.

\subsection{Understanding FMs and SMs}
\textbf{Open-set Capability.}
We split the dataset into training and test datasets with no overlapped classes to test the open-set capability of FMs and small models (SMs).
Table~\ref{tab:moti_general} shows that FMs can achieve up to 77\% mean accuracy without any fine-tuning or calibration, while SMs only show 1.5\% accuracy on average, which is equivalent to random guessing.
The significant performance gap shows that FMs can be used for open-set capability while SMs are hardly usable when the test data is not seen during the training. 

\setlength{\tabcolsep}{4pt}
\begin{table}
  \caption{The performance of small models (SMs) and foundation model (FMs) with unseen test samples on FLO102 and SC40 datasets, respectively.}
  \label{tab:moti_general}
  \begin{tabular}{@{}cccccccccl@{}}
    \toprule
 &Models& FLO102  & SC40 & Param. & FLOPS  & Nano\\
\midrule
\multirow{2}{*}{SMs} & MobileNet & 1.1\%   & 2.6\% & 3.5M & 0.3B &36.8ms\\
& ResNet18 & 0.4\%  & 3.4\% & 11.7M & 1.8B &30.5ms\\
\midrule
\multirow{2}{*}{FMs}&  ImageBind& 78.4\% &  71.3\% &  1172M &  167.3B  &N.A. \\
& CLIP-L/14& 79.5\% &  77.1\% &  407.8M  & 61.5B & N.A. \\
  \bottomrule
\end{tabular}
\end{table}



\begin{figure}
\begin{minipage}[t]{0.49\columnwidth}
     \centering
\includegraphics[width=1\textwidth]{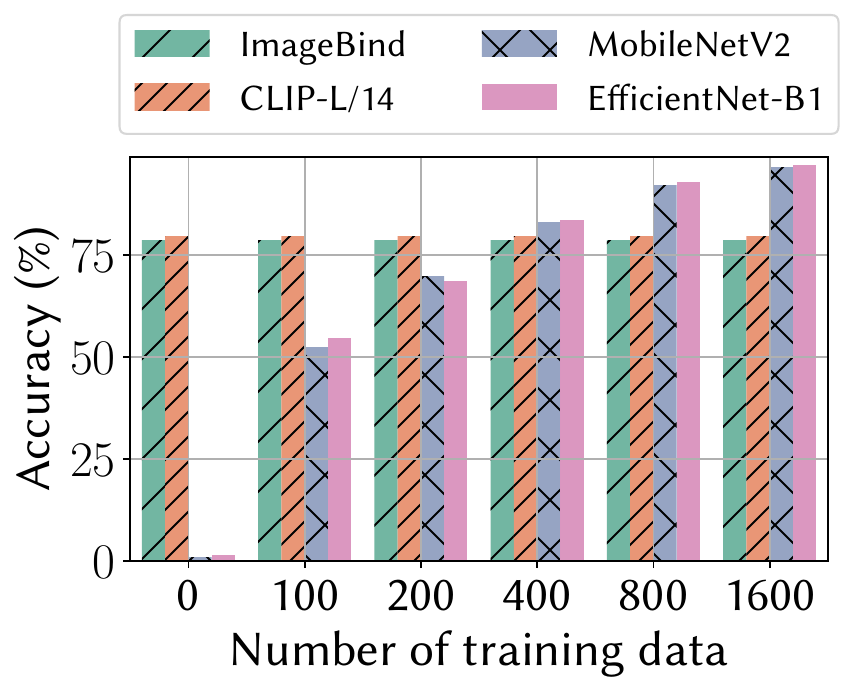}
        \caption{The performance of SMs with different amounts of fine-tuned data.
        The parameters of FMs are frozen.
        }
        \label{fig:motivation}
\end{minipage}
\hfill
   \begin{minipage}[t]{0.49\columnwidth}
     \centering
       \includegraphics[width=1\textwidth]{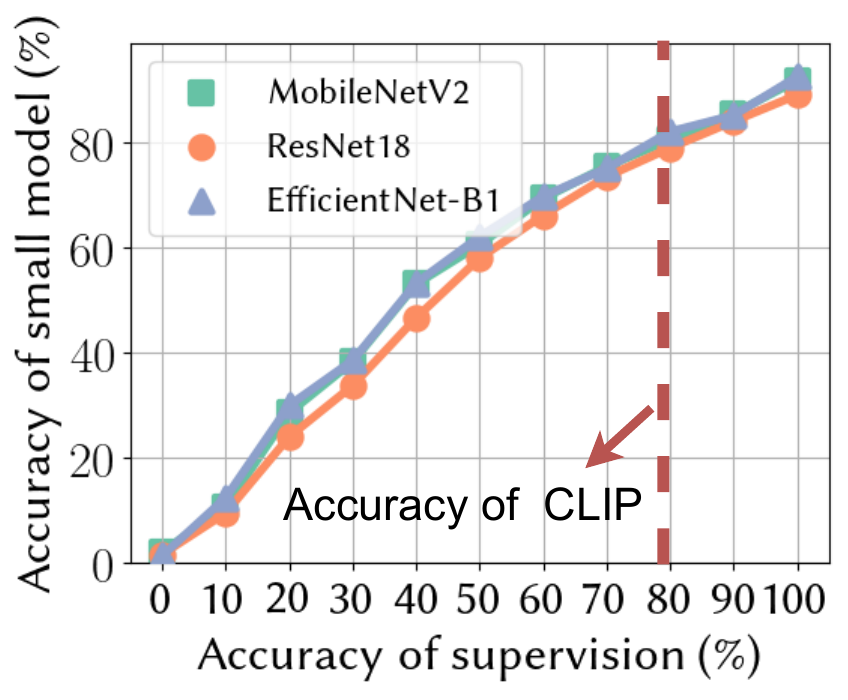}
    \caption{Feasibility of leveraging FMs as a knowledge base for customizing small models.}
    \label{fig:motivation_pseudo}
\end{minipage}
\end{figure}

\noindent\textbf{Customization of Small Models.}
We further test the performance of customized small models by calibrating them with data from the new classes (i.e., the new types of objects added to the environment).
Figure~\ref{fig:motivation} shows the accuracy of the small models customized by different amounts of labeled data from the new classes.
SMs show unsatisfactory performance that is far below that of FMs when the amount of labeled data is limited. When more labeled data is used for calibration, SMs can achieve a good performance that is even higher than FMs, up to 92\%.
This shows that the fully customized small model can achieve similar and even superior recognition performance than FMs on specific tasks.



    
    


\noindent\textbf{Execution Latency of FMs and SMs.}
Table~\ref{tab:moti_general} shows the parameter size and inference latency of customized SMs and FMs.
The parameter size and computational overhead in FLOPS of MobileNetV2 are 335x and 558x less than ImageBind, respectively.
The inference latency for one single image of MobileNetV2 and ResNet18 are 36.8ms and 30.5ms on Jetson Nano, respectively. 
However, both ImageBind and CLIP-L/14 require more than 6GB memory, which exceeds the memory limit of Jetson Nano, and thus can not directly execute (marked as N.A. in Table~\ref{tab:moti_general}).
The huge computation requirement of FMs urges it mostly deployed on the cloud.

The above preliminary results motivate us to adopt a model-switching solution between FMs on the cloud and customized small models on the edge, which can leverage the respective advantages of FMs and customized small models. 
In summary, FMs like CLIP and ImageBind have good open-set capabilities, but they are too large to fit in embedded systems. On the other side, SMs can achieve a better performance than FMs after customizing models with labeled data from the new classes. However, the labeled data for customization is usually not available in practical applications.






\subsection{Customization with FMs}
We further explore the feasibility of leveraging FMs to customize SMs by using the prediction results of FMs as pseudo labels to supervise the customization of SMs.
Specifically, we train SMs under different percentages of correct labels and validate whether training with noisy labels is possible.
Figure~\ref{fig:motivation_pseudo} shows the performance of SMs when we set different accuracy of the pseudo labels. We control the correct percentage of the pseudo labels among 400 unseen test samples, which is the values on the x-axis.
The vertical line shows the accuracy of CLIP-L/14.
The result shows that the accuracy of the SM is 90\% when all the pseudo labels are correct, while the accuracy of the SM drops to 80\% when the pseudo labels have 80\% accuracy. 
Meanwhile, CLIP can provide accuracy with 79.5\% without any fine-tuning.
This motivates us to use FMs as a rich and general knowledge base to provide high-quality supervision for the customization of SMs.

\section{Application and System Overview}
EdgeFM is a novel edge-cloud cooperative system with open-set recognition ability by using FMs for selective knowledge query and edge model customization.
Next, we will introduce the application scenario and overview of the system.

\subsection{Application Scenarios}
We use two examples to discuss the potential applications of EdgeFM.
The first application is robot semantic sensing \cite{meyer2019importance} which is widely used for home services \cite{kishida2021object} and industrial applications \cite{vallachira2019data}. 
The main challenge for the current home-service robot system is the high diversity in personal items and dynamic environmental factors across households.
In addition, the types of objects in the environment will change over time. For example, users sometimes buy or throw items.
Current recognition models on home-service robots \cite{meyer2019importance,kishida2021object} require labeled data for model retraining to adapt to the new classes of objects, which is not realistic and user-unfriendly.
The second application is human activity recognition (HAR), which is an essential algorithm for many smart embedded systems for health, such as disease monitoring and fitness tracking.
The main challenge for current HAR systems \cite{ouyang2022cosmo} is that there is a need for \textit{model calibration}.
During the first-time installation, the HAR system requires the user to manually collect and label data to calibrate the model to the environment and the target activities of users.
Moreover, it may require periodic calibration when the monitored activities change due to the changing health condition or the doctor's advice.

\subsection{System Architecture} 
\begin{figure*}[tb]
  \centering
\includegraphics[width=0.98\linewidth]{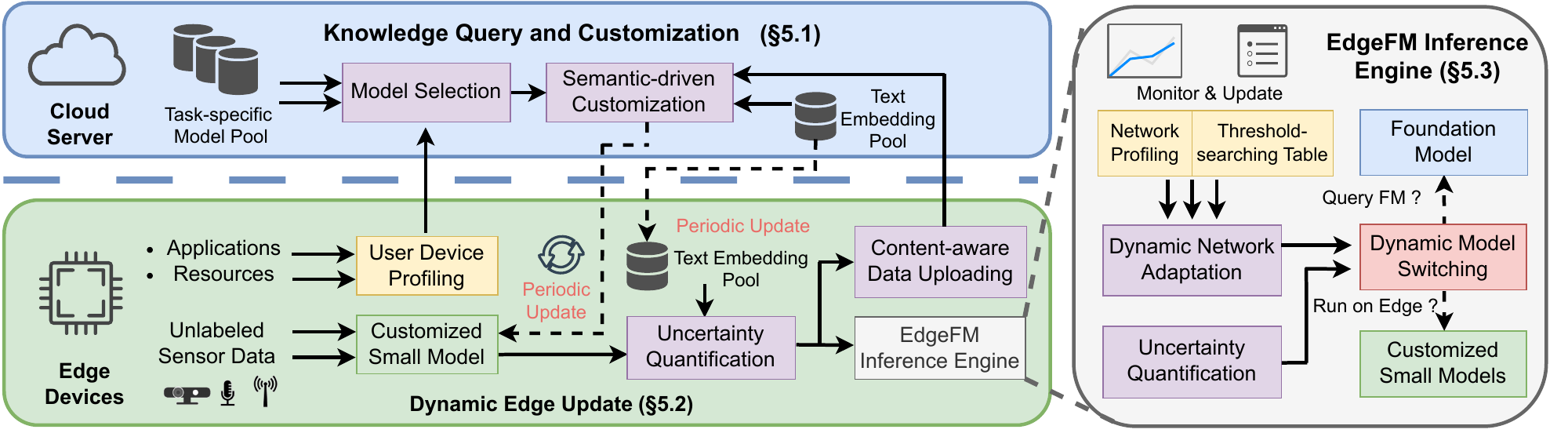}
  \caption{Overall system architecture of EdgeFM.}
  \label{fig:COCLIPdesign}
\end{figure*}

EdgeFM is a novel edge-cloud cooperative system that enables edge devices with open-set recognition.
Figure~\ref{fig:COCLIPdesign} shows the system overview of EdgeFM.
In the customization stage, EdgeFM conducts \textit{knowledge query and customization} (\S~\ref{Knowledge Query and Customization}) and \textit{dynamic edge update} (\S~\ref{Dynamic edge update}) to selectively upload unlabeled data to the cloud and customize the domain-specific knowledge and architectures for small models.
Specifically, EdgeFM first conducts user device profiling on the edge to obtain information about the applications (e.g., tasks and modalities) and computation resources (e.g., memory constraints) of edge devices (\S~\ref{User Device Profiling}).
The profiling results are then employed by the model selection module to determine the appropriate architecture for small models on the edge (\S~\ref{Architecture of Customized Small Models}).
Meanwhile, EdgeFM selectively uploads the unlabeled sensor data to the cloud (\S~\ref{Data uploading}) for customization.
EdgeFM will conduct semantic-driven customization on the cloud  (\S~\ref{customization}) and periodically update the customized small model and text embedding pool to the edge.
During the inference, EdgeFM inference engine employs the dynamic network adaptation module (\S~\ref{Dynamic Network Adaptation}) to continuously monitor the network condition and update the threshold searching table.
Then, EdgeFM conducts dynamic model switching (\S~\ref{Dynamic Model Switching}) to determine whether query FM on the cloud or use customized small models for inference.
The dynamic model switching policy of EdgeFM supports open-set models and also considers both the uncertainty of the sensor data and the dynamic network variation.

\section{Design of EdgeFM}

\subsection{Knowledge Query and Customization}
\label{Knowledge Query and Customization}
In this section, we will introduce how EdgeFM customizes the domain-specific knowledge and architectures for the small models in a label-free manner.






\subsubsection{Semantic-driven Customization}
\label{customization}
The limited computation resources of edge platforms and real-time requirements of tasks usually require adopting mobile-friendly CNNs architectures rather than the heavyweight transformer \cite{vaswani2017attention}.
The main challenge here is how to effectively customize the heterogeneous lightweight CNNs by the knowledge from FMs in a \textit{label-free manner} while preserving open-set recognition capability.
To address this challenge, we propose a semantic-driven customization approach.
Next, we will introduce the components of our semantic-driven customization approach respectively.






\noindent\textbf{Heterogeneous Feature Mapping.}
Unlike existing work distilling knowledge between similar architectures, EdgeFM conducts customization between heterogeneous models, i.e. FMs to mobile-friendly CNNs.
Existing lightweight CNNs \cite{howard2017mobilenets,tan2019efficientnet} can be regarded as consisting of a convolution-based feature extractor and a task-specific classifier.
However, FMs usually adopt transformer-based architectures \cite{vaswani2017attention}, which encode the input images or spectrograms into a sequence of tokens and extract context information.
Multi-modal FMs further align the embeddings of vision or audio modality with the text embedding in a unified embedding space.
The difference in embedding space and heterogeneous model architectures between FMs and small models makes the customization challenging.
Therefore, we discard the task-specific classifier of the original small models and add a feature projection network on top of the original feature extractor of small models, which is defined as $\mathbf{v}_{i} = \psi (\mathcal{S}(\mathbf{x}_i))$, where $\mathcal{S}(\cdot)$ denotes the feature extractor of the customized small model, $\psi(\cdot)$ is the feature projection network. 
The architecture of the feature projection network $\psi(.)$ is a lightweight single-layer feed-forward network. 
It can ensure the features of the customized small model have the same dimension as FM's unified embedding space, facilitating matching with text embeddings of FMs to enable open-set recognition.
For example, the output features of MobileNetV2 have a dimension of 1280, while the unified embedding space for ImageBind is 1024. Consequently, the input and output dimensions of the feature projection network are 1280 and 1024, respectively.



\begin{figure}
    \centering
    \begin{subfigure}{0.49\columnwidth}
        \centering
\includegraphics[width=0.99\textwidth]{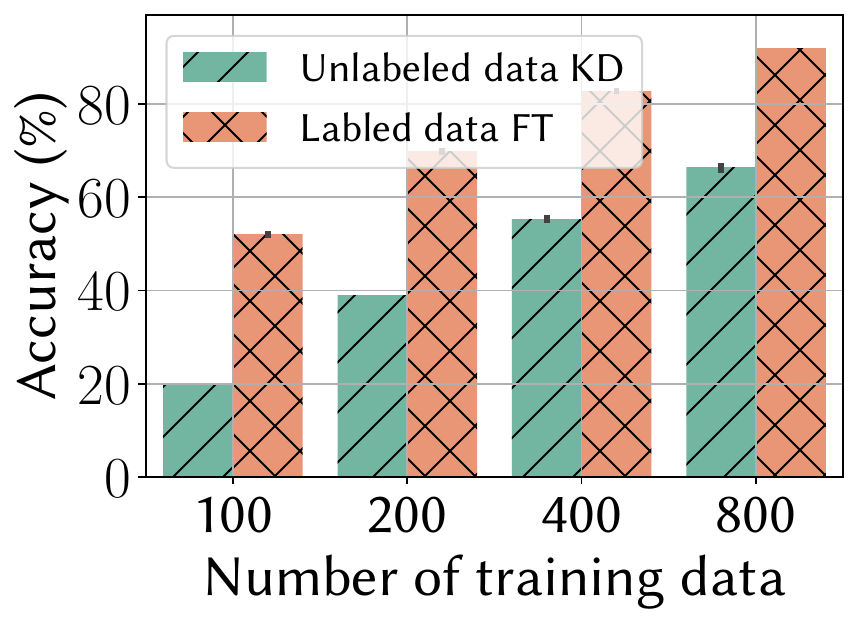}
\caption{Flower recognition.}
\label{fig:KD_motivation_flo}
    \end{subfigure}
    \hfill
    \begin{subfigure}{0.49\columnwidth}  
     \centering 
\includegraphics[width=0.99\textwidth]{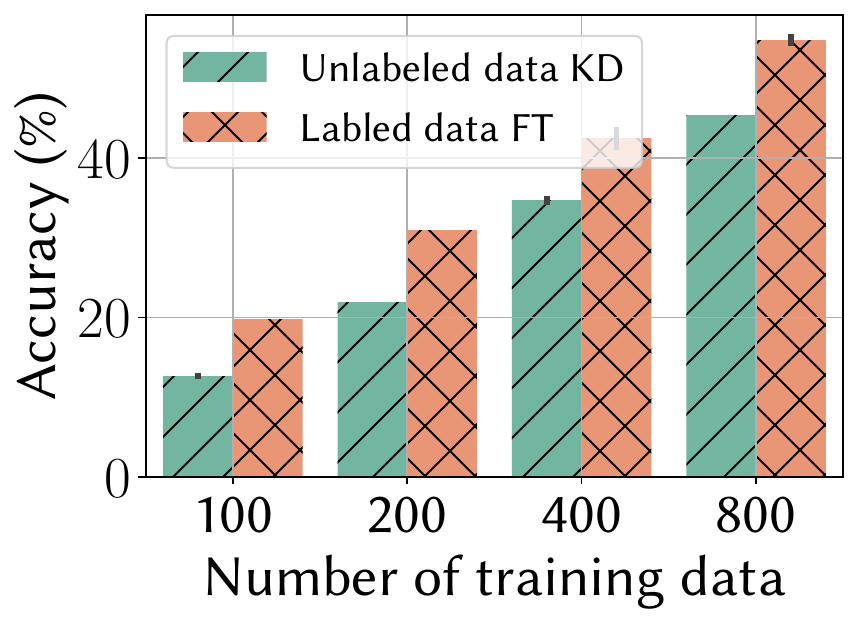}
\caption{Activity recognition.}
\label{fig:KD_motivation_ucf}
    \end{subfigure}
  \caption{Customization accuracy with labeled data and unlabeled data. FT denotes fine-tuing.
  } 
  \label{fig:KD_motivation}
\end{figure}

\noindent\textbf{Knowledge Query from the Foundation Model.}
It is impractical to obtain high-quality labeled data in embedded systems.
Under this scenario, the direct way to customize domain-specific knowledge from FMs is to use Mean-Squared-Error (MSE) loss to pull closer the embedding of sensor modality (e.g., image) between FMs and small models.
Figure~\ref{fig:KD_motivation} shows the recognition accuracy of the customized small model when fine-tuned with labeled data and when employing unlabeled data with MSE loss for knowledge distillation from FM.
We can see that employing unlabeled data with MSE loss will lead to significant accuracy degradation compared with using labeled data for fine-tuning.


To address this challenge, we propose to leverage FMs to further customize the \textit{user-specific knowledge} to the small models.
Here we define the user-specific knowledge as the interested class set specified by users.
Specifically, EdgeFM initially pre-stores a \textit{text embedding pool} that contains text embeddings from a wide range of frequently-used classes.
The text embeddings in the pool are computed by the text encoder of FMs on the cloud.
In practical use, EdgeFM allows users to freely add their interested classes for respective applications.
The text embeddings of these newly added classes are computed by FMs on the cloud and are added to the pool, which is then updated to the user device periodically (\S~\ref{User Device Profiling}).
Note that it does not mean requiring users to annotate each data, but only providing interested classes set.

Take vision tasks as an example, suppose the text embedding pool is $ \mathbf{T} $ and the visual encoder of FM is $ \mathcal{T}_v(\cdot) $, FMs on the cloud first extract the visual embedding of the unlabeled data $\mathbf{x}_i$ as $\mathcal{T}_v(\mathbf{x}_i)$.
Then, EdgeFM will select a text embedding $\mathbf{t}_i^\prime$ from the text embedding pool with the highest similarity with $\mathcal{T}_v(\mathbf{x}_i)$ as:
\begin{equation}
    \mathbf{t}_i^\prime =arg max (\left \langle \mathcal{T}_v(\mathbf{x}_i),\mathbf{t}_k \right \rangle ) 
    , \mathbf{t}_k\in \mathbf{T}, 
    \label{pseudolabel_1}
\end{equation}
where $\mathbf{t}_i^\prime$ is defined as the pseudo text embedding. 
We also assign $\mathbf{t}_i^\prime$ a confidence score as $w_i = \left \langle \mathcal{T}_v(\mathbf{x}_i),\mathbf{t}_i^\prime \right \rangle $, i.e. the cosine similarity between $\mathcal{T}_v(\mathbf{x}_i)$ and $\mathbf{t}_i^\prime$.
The obtained pseudo text embedding and its confidence score will be used for semantic-driven distillation.

\noindent\textbf{Semantic-driven Distillation Loss.}
To further customize the user-specific knowledge, we propose semantic distillation loss.
Here we take vision recognition as an example.
Firstly, we adopt MSE loss to pull the features extracted by customized small models closer to the visual embedding of FMs, i.e.
$\mathcal{L}_{vis} = \mathcal{H}_{MSE}(\mathcal{T}_v(\mathbf{x}_i),\mathbf{v}_{i} )$,
where $ \mathcal{T}_v(\mathbf{x}_i) $ and $ \mathbf{v}_{i} $ are the visual embedding of FMs and customized small model, respectively;
$\mathcal{H}_{MSE}$ is MSE function.
Next, we adopt a bidirectional contrastive learning loss \cite{radford2021learning} to further pull the features extracted by customized small models closer to the pseudo text embedding of FMs.
Given a mini-batch of $ bs $ paired data ($\mathbf{v}_i$, $ \hat{\mathbf{t}}_{i} $), where $\mathbf{v}_i$ is the embedding of the sensor data (e.g., image) extracted by the customized small model, and $ \hat{\mathbf{t}}_{i} $ is the most similar text embedding in the text embedding pool of FMs, as in Equation~\ref{pseudolabel_1}.
We adopt the bidirectional contrastive learning loss which is defined as:
\begin{equation}
  \mathcal{L}_{i}^{v \rightarrow t^\prime} = 
  -log\frac{exp	\left\{ \left \langle \mathbf{v}_{i}, \hat{\mathbf{t}}_{k}  \right \rangle / \tau  	\right\}  }
  {\sum_{k=1}^{bs}
  exp\left\{ \left \langle \mathbf{v}_{i}, \hat{\mathbf{t}}_{k} \right \rangle / \tau  \right\} }
\end{equation}

\begin{equation}
  \mathcal{L}_{i}^{t^\prime \rightarrow v} = 
  -log\frac{exp\left\{ \left \langle \hat{\mathbf{t}}_{i}, \mathbf{v}_{k}  \right \rangle / \tau  \right\}  }
  {\sum_{k=1}^{bs}
  exp\left\{ \left \langle \hat{\mathbf{t}}_{i}, \mathbf{v}_{k} \right \rangle / \tau  \right\} }
\end{equation}

\begin{equation}
  \mathcal{L}_{text} = \frac{1}{bs}\sum_{i=1}^{bs}
  w_i 
  \left\{
  \lambda  \mathcal{L}_{i}^{v \rightarrow t^{\prime}} + (1-\lambda) 
  \mathcal{L}_{i}^{t^{\prime} \rightarrow v} 
  \right\}
\end{equation}
where $ w_i $ is the confidence score of the sample $\mathbf{x}_i$, which is obtained as mentioned before.
$\lambda$ and $\tau$ are the weight and temperature parameters, respectively.
We test the two parameters with FLO102 dataset and choose $\lambda=0.5$ and $\tau=1$ that achieve the best recognition performance for the customized small model.

\subsubsection{Model Selection of Small Models}
\label{Architecture of Customized Small Models}

\begin{figure}
    \centering
    \begin{subfigure}{0.49\columnwidth}
        \centering
        \includegraphics[width=\textwidth]{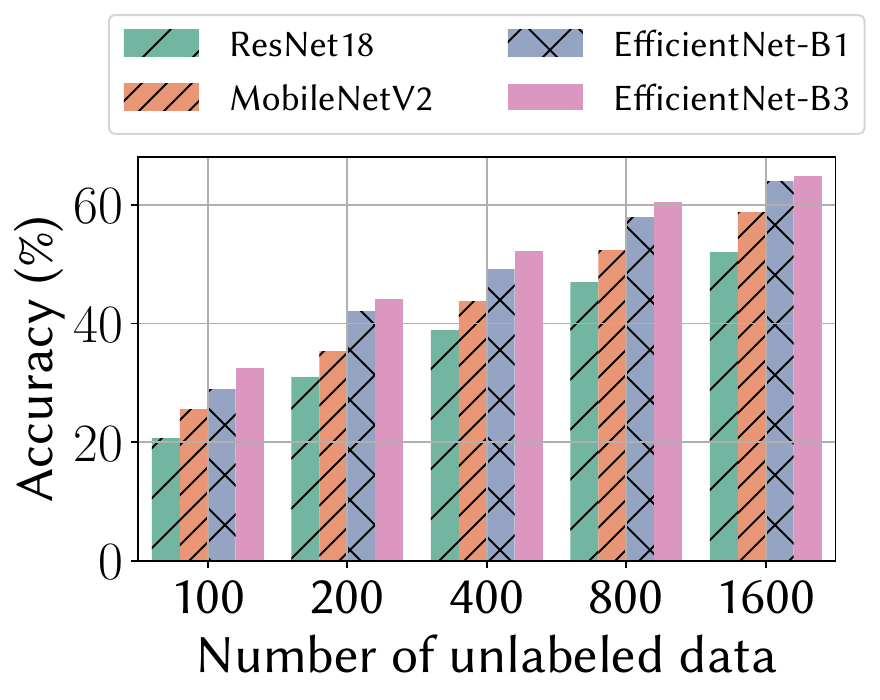}
        \caption{Activity recognition.}  
        \label{fig:smallmodel_ucf}
    \end{subfigure}
    \hfill
    \begin{subfigure}{0.49\columnwidth}  
        \centering 
        \includegraphics[width=\textwidth]{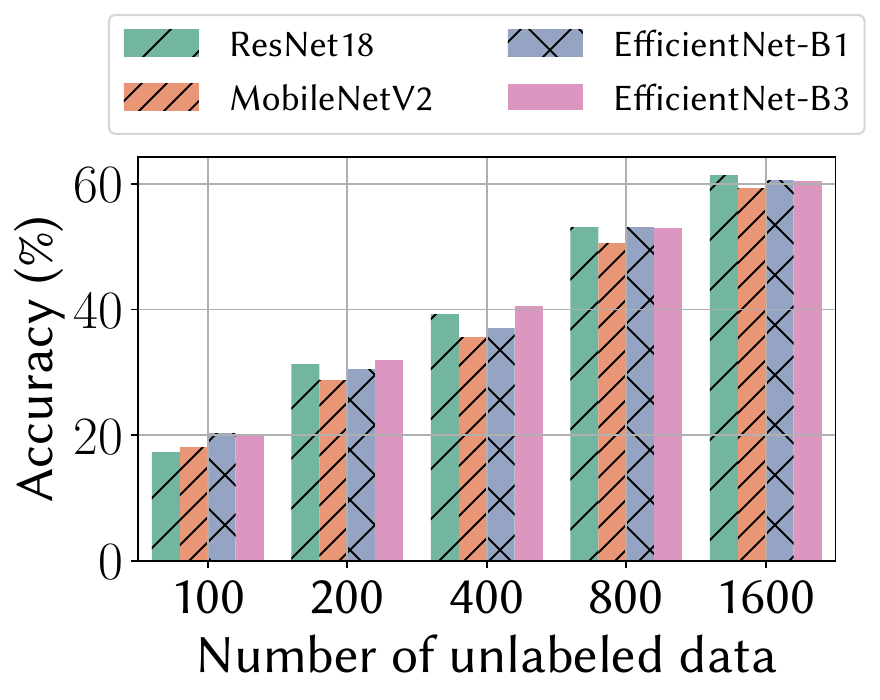}
        \caption{Audio recognition.}    
        \label{fig:smallmodel_esc}
    \end{subfigure}
    
  \caption{
  Performance of small models with different architectures on different modalities.
  }
  \label{fig:smallmodel_arch}
    \vspace{-1.0em}

\end{figure}

We develop a model selection module that can customize the architecture of small models on the edge based on the tasks, modalities, and computation resources of edge devices.
Figure~\ref{fig:smallmodel_arch} shows varied performance of four small models with different architectures on different tasks and data modalities.
In particular, MobileNetV2 \cite{howard2019searching} performs better on vision-based recognition tasks like HAR, but has worse performance on the audio recognition task.
This is because the depth-wise separable convolution is unsuitable for extracting the features from spectrogram-based data \cite{howard2017mobilenets}.
Moreover, the computation resources such as memory footprint and FLOPS vary for different edge devices and tasks.
To this end, EdgeFM will determine the architecture of small models based on the tasks, data modalities, and computation resources of edge devices.
Specifically, EdgeFM pre-stores many classical small model architectures with different accuracy and computation overhead such as MobileNet \cite{howard2017mobilenets}, and also those optimal architectures searched by Neural Architecture Search (NAS) technique such as EfficientNet series \cite{tan2019efficientnet}, on the cloud server.
These small model architectures are grouped and stored in a task-specific model pool according to the modalities and tasks.
At the offline stage, we test the recognition accuracy of each small model on public datasets and measure the resource usage, including the memory footprint and FLOPS.
Note that we only use the model's accuracy on public datasets to assess its representation capability, without necessitating the use of labeled data from users.
The accuracy, memory footprint, and FLOPS are recorded in a table, i.e., the accuracy-resource lookup table.
At the online stage, EdgeFM will first select the corresponding model pool $POOL_{app}$ based on the application specified by users.
Next, EdgeFM determines the architecture of the small model by searching the accuracy-resource lookup table to maximize the recognition accuracy under the resource constraints of FLOPS and memory.





\subsection{Dynamic Edge Update}
\label{Dynamic edge update}
\subsubsection{Content-aware Data Uploading}
\label{Data uploading}
Streaming all data to the cloud can cause non-trivial transmission overhead and unreliability.
However, existing approaches \cite{li2021appealnet,park2015big} utilize the softmax score by the conventional closed-set model to determine the data uploading, which is not suitable for open-set models. 
Therefore, we design a content-aware data-uploading approach tailored for open-set models, which utilizes the semantic similarity between sensor data embeddings and text embeddings as uploading criterion.



Since our semantic-driven customization enables the open-set recognition capability of the customized small models, we use semantic similarity as uncertainty quantification of the samples.
This distinguishes EdgeFM from previous studies in terms of the uncertainty metrics.
Specifically, for each collected data samples $\mathbf{x}_i$, we first compute the cosine similarity between sensor data embedding $\mathbf{v}_{i}$ (computed by the customized small models) and each text embedding $\mathbf{t}_{k}$ in the text embedding pool $ \mathbf{T} $, which is defined as $sim(\mathbf{x}_i)=\left \langle \mathbf{v}_{i}, \mathbf{t}_{k}  \right \rangle$.
We use the margin score \cite{park2015big} as uncertainty quantification for the samples, which is defined as $Unc(\mathbf{x}_i) = sim_1(\mathbf{x}_i) - sim_2(\mathbf{x}_i) $,
where $sim_1(\mathbf{x}_i)$ and $sim_2(\mathbf{x}_i)$ are the highest similarity and the second highest similarity calculated between $\mathbf{x}_i$ and all $\mathbf{t}_{k}$ in the text embedding pool.
Only samples with $ Unc(\mathbf{x}_i) < V_{thre} $ will be uploaded to the cloud for customization.
We set $V_{thre}=0.99$ based on the observation that this configuration results in a substantial reduction in data transmission with negligible accuracy deterioration.

\begin{figure}
    \centering
    \begin{subfigure}{0.495\columnwidth}
        \centering
\includegraphics[width=0.995\textwidth]{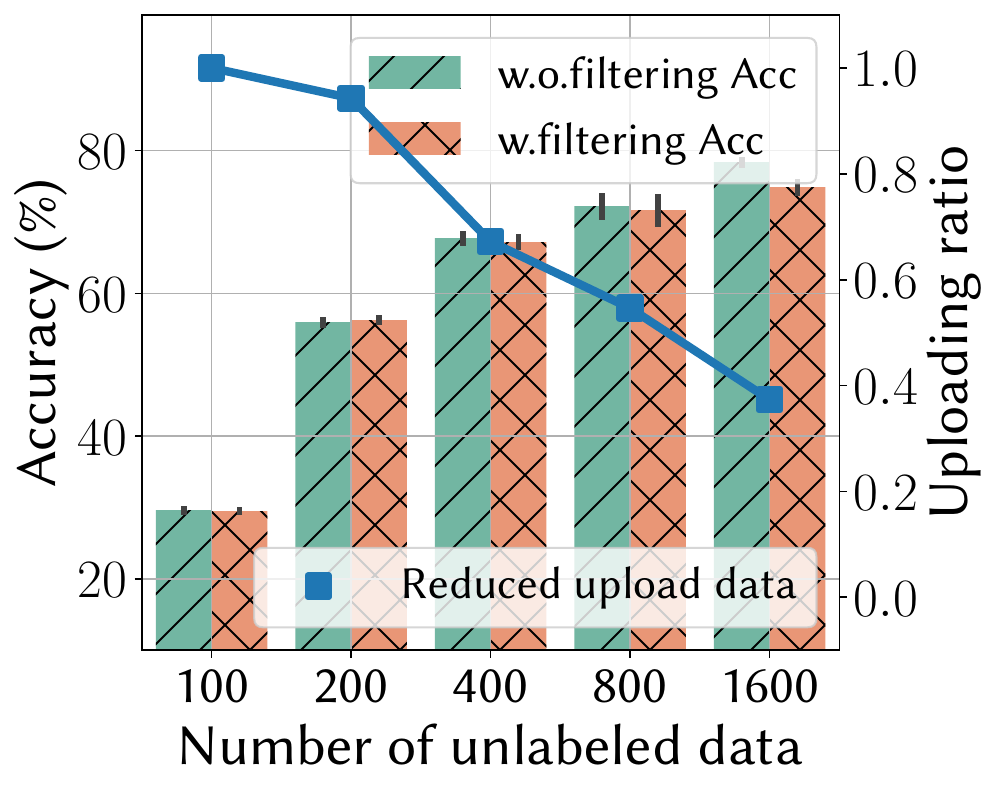}
\caption{Flower recognition.}
\label{fig:KD_filtering_flo}
    \end{subfigure}
    \hfill
    \begin{subfigure}{0.495\columnwidth}  
     \centering 
\includegraphics[width=0.995\textwidth]{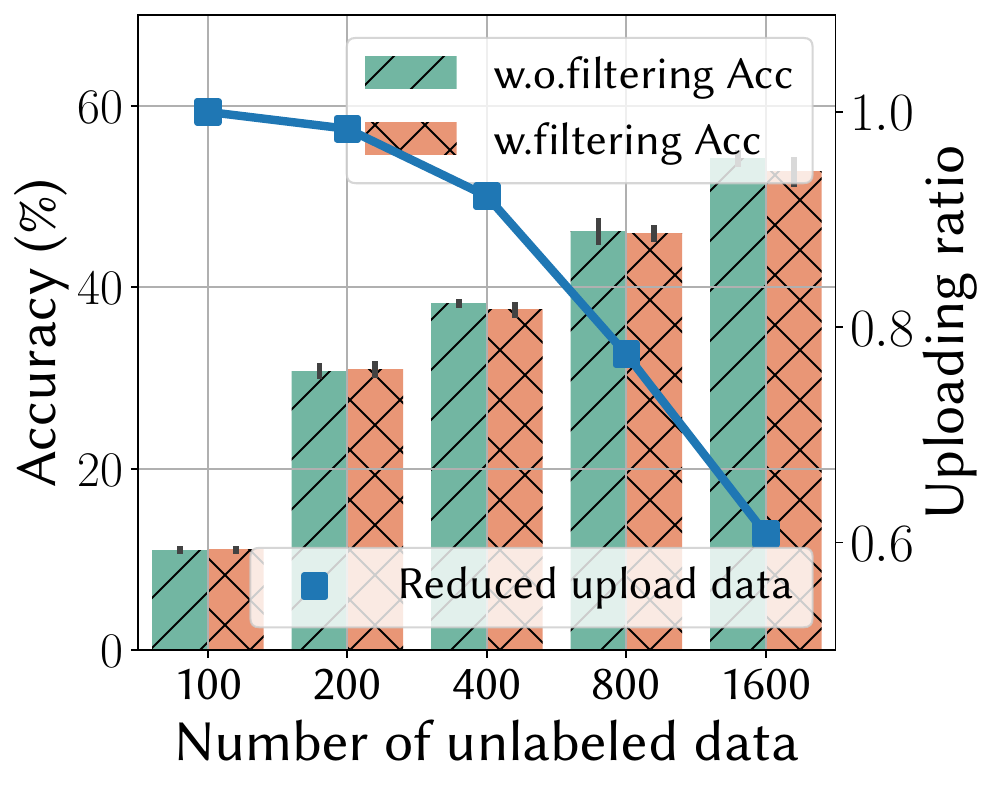}
\caption{Activity recognition.}
\label{fig:KD_filtering_ucf}
    \end{subfigure}
    
  \caption{
  Customization accuracy and uploading data ratio with or without content-aware data-uploading. 
  }
  \label{fig:KD_filtering}
  \vspace{-1em}
\end{figure}

Figure~\ref{fig:KD_filtering} shows the customization accuracy and uploading data ratio with or without content-aware data uploading.
We can see that with the collected unlabeled sensor data increasing from 100 to 1600, the ratio of uploading data decreases from 100\% to about 40\% on two applications (as the blue line shows) with a negligible accuracy drop.
Therefore, content-aware data uploading can help EdgeFM reduce the network transmission overhead in real-world implementations.

\subsubsection{User Device Profiling and Periodic Update}
\label{User Device Profiling}
To reduce the transmission overhead, EdgeFM periodically updates the customized small model and text embedding pool and delivers them to the edge device.
Compared to prior studies \cite{bhardwaj2022ekya,khani2023recl}, a unique characteristic of our approach is the dynamic updating of the text embedding pool on the edge side, which enables the support of open-set recognition on edge devices.
Specifically, EdgeFM utilizes a user device profiler to record the information of edge devices, such as applications (e.g., tasks and modalities), and computation resources of edge devices (e.g., memory usage and latency requirements).
This information is then employed by the model selection module (\S~\ref{Architecture of Customized Small Models}) to select the appropriate architecture for the customized small model. 
On the other hand, EdgeFM continuously collects sensor data from the environment and selectively uploads them to the cloud server.
Upon the uploaded data reaching the specified amount, EdgeFM will conduct semantic-driven customization and subsequently download the updated customized small model to the edge device.
Moreover, the text embedding pool on the cloud is also updated if users add their interested class set.
The text embedding pool and customized small model will be updated synchronously to the user device.
The frequency of periodically updating the edge side in EdgeFM offers a trade-off between accuracy and transmission overhead. 
Since the experimental results in \cite{bhardwaj2022ekya} have shown that setting the updating interval to $200\,\text{sec}$ yields the best trade-off between accuracy and transmission overhead, EdgeFM adopts the same updating interval of $200\,\text{sec}$ for both customized small models and the text embedding pool.

\subsection{EdgeFM Inference Engine}
This section introduces the EdgeFM inference engine, which performs dynamic model switching at runtime, considering both the uncertainty of sensor data and network variation.


\subsubsection{Dynamic Model Switching}
\label{Dynamic Model Switching}
EdgeFM adopts an edge-cloud hybrid prediction mechanism
based on the collaboration between customized small models, router model, and FM, where the first two models run on edge devices while FM runs on the cloud.
The overall prediction of EdgeFM for input sample $\mathbf{x}_i$ is defined as:
\begin{equation}
     P(\hat{y} \mid \mathbf{x}_i) = 
     r(\mathbf{x}_i)P_{SM}(\hat{y} \mid\mathbf{x}_i)
     + (1-r(\mathbf{x}_i))P_{FM}(\hat{y} \mid\mathbf{x}_i)
\end{equation}
where $P_{SM}(\hat{y}\mid\mathbf{x}_i)$ and $P_{FM}(\hat{y}\mid\mathbf{x}_i)$ are the predictions of the customized small model and FM, respectively.
Note that the predictions of the open-set model in EdgeFM are computed by the cosine similarity score between sensor data embeddings and text embeddings, i.e. $P_{SM}(\hat{y} 
\mid\mathbf{x}_i)=\left \langle \mathbf{v}_{i}, \mathbf{t}_{k}  \right \rangle$,
$P_{FM}(\hat{y} \mid\mathbf{x}_i)=\left \langle \mathcal{T}_v(\mathbf{x}_i),\mathbf{t}_{k} \right \rangle$.
$\mathbf{v}_{i}$ and $\mathcal{T}_v(\mathbf{x}_i)$ are the sensor data embeddings that are computed by the customized small model and FM, respectively.
A router model $r(\mathbf{x}_i)$ controls the models switching according to the prediction of the customized small model and a threshold $thre (t)$, which is defined as:
\begin{equation}
     r(\mathbf{x}_i) = \mathds{1}\left\{
     Unc(\mathbf{x}_i) \geq thre(t)
     \right\}
\end{equation}
where $Unc(\mathbf{x}_i)$ is the uncertainty of the sensor data, which is defined in Section~\ref{Data uploading}.
Note that the threshold $thre (t)$ in the inference engine is different from the threshold in the dynamic edge update. 
EdgeFM tunes the threshold $thre (t)$ at runtime to adapt to the dynamic network condition (see Section~\ref{Dynamic Network Adaptation}).
Based on the uncertainty of the input sample and threshold, the router model determines whether to query FMs on the cloud for inference or use the prediction of the customized small model.

\subsubsection{Dynamic Network Adaptation}
\label{Dynamic Network Adaptation}
Model switching threshold determines the trade-off between accuracy and inference latency.
EdgeFM adopts the dynamic network adaptation module to find the optimal model switching threshold under the dynamic network fluctuation.


Specifically, EdgeFM will periodically collect a specific number of sensor data from the environment to build a calibration set.
We sample the threshold equally in the range of $(0,1)$.
For each $thre \in (0,1)$, EdgeFM computes the edge-side processing proportion $r(thre)$, overall accuracy $acc(thre)$, edge-side processing latency $t_{edge}$, transmission latency $t_{trans}$, and cloud-side processing latency $t_{cloud}$ for the calibration set, and saves them in a threshold-searching table.
The estimated end-to-end inference latency can be expressed as:
\begin{equation}
\hat{t}_{e2e}(thre)  = r(thre)\cdot t_{edge} + (1-r(thre))\cdot(t_{trans}+t_{cloud})
\label{eq:1}
\end{equation}
As EdgeFM does not need users to provide data annotations, we adopt the predictions of FM as ground truth to compute the accuracy as the estimated accuracy $acc(thre)$.

At runtime, EdgeFM performs analysis on the estimated accuracy-latency space based on the threshold-searching table and the priority of user demands.
For example, if the priority of accuracy is higher than the inference latency, EdgeFM will select the smallest $thre$ to satisfy not exceeding the constraint of accuracy degradation.
If the inference latency has a higher priority, EdgeFM will select the largest $thre$ to ensure the estimated end-to-end latency is lower than the latency constraint as follows:
\begin{equation}
\max_{thre\in(0,1)} \ thre \quad
\textrm{s.t.} \quad 
\hat{t}_{e2e}(thre) \leq L_{app}
\label{eq:2}
\end{equation}
where $L_{app}$ is the end-to-end latency constraint, which is specified by the applications.
At runtime, network transmission time $t_{trans}$ can be efficiently updated by estimating the real-time network bandwidth: $t_{trans} = \frac{Dim}{B(t)}$, where $Dim$ is the dimension of samples, $B(t)$ is the real-time estimated network bandwidth.
The estimation of the network has been extensively studied in prior research \cite{kim2001mobile}.
Our approach is compatible with the most widely used techniques in this field.
Based on Equation~\ref{eq:1} and Equation~\ref{eq:2}, we can obtain the optimal edge-side processing proportion $r(thre)$ at the current network bandwidth.
The relationship between $r(thre)$ and $thre$ can be queried from the threshold-searching table with negligible latency.
Therefore, EdgeFM can adapt to the network variation through dynamic adjusting its threshold at runtime.


\subsection{System Implementation}

\subsubsection{Edge-cloud Implementation}
We implement EdgeFM on a desktop server (Intel i9-12900K CPU with two NVIDIA RTX 3090 GPU) and two NVIDIA edge platforms, Jetson Nano and Jetson AGX Xavier.
The network connection and data transmission parts in EdgeFM are developed via TCP socket API.
We use Traffic Control in Linux to simulate different network conditions and iPerf tool \cite{iperf} to measure the network bandwidth at regular one-second intervals.

    

\subsubsection{Foundation Models}
We implement EdgeFM on two FMs, ImageBind \cite{girdhar2023imagebind} and CLIP \cite{radford2021learning}.
We adopt the image and audio modalities of ImageBind for evaluation in this work.
For CLIP, we use the CLIP-L/14 version, 
which reports the highest performance among CLIP series.
For vision-based tasks, we use CLIP and the vision branch of ImageBind as FMs for evaluation.
As CLIP only supports vision modality, we only use ImageBind for the audio recognition task.

\subsubsection{Prompt Setting}
\label{Prompt_setting}
Both ImageBind and CLIP require a prompt to convert the single class name in a natural language manner into a textual description. 
The prompt for HAR task is set to ``a photo of a person doing {$CLS$}.'', which are the same as CLIP's setting \cite{radford2021learning}.
For indoor scene recognition and flower recognition, the prompt is ``a photo of a {$CLS$}.''.
For audio recognition, we extract the text embeddings from the class name, which is the same as ImageBind's setting \cite{girdhar2023imagebind}.

\subsubsection{Baseline Approaches}
We compare EdgeFM with two types of baselines, including the 
efficient on-device inference baselines and open-set recognition baselines. 

\noindent\textbf{Efficient On-device Inference Baselines.}
We implement several representative on-device NN efficient inference approaches on ImageBind and CLIP for a fair evaluation. 

\textit{PersEPhonEE} \cite{leontiadis2021s}, which is an edge-only NN acceleration approach based on early exit.
we implement PersEPhonEE on the two FMs.
There are two ways to implement the early-exit classifier for ImageBind and CLIP, including a fully-connected classifier and cosine distance classifier \cite{chen2019closer}, where the latter one is adopted as it performs better in experiments.


\textit{SPINN}
\cite{laskaridis2020spinn}, which is an edge-cloud collaboration approach integrating model splitting and early-exit techniques.
Similarly, we re-implement SPINN on ImageBind and CLIP.
We also adopt cosine distance classifier as the early-exit head.

\textit{Cloud-centric}, which is the most widely adopted solution for FM inference \cite{liang2022effective,shah2022lm,ahn2022can}.
For the cloud-centric approach, we deploy ImageBind and CLIP on the server and offload all the samples to the server for inference.

\noindent\textbf{Open-set Recognition Baselines.}
We also compare the open-set recognition accuracy of EdgeFM with other open-set recognition baselines.

\textit{Semantic-based Approaches.} 
DUS-VAE \cite{su2022distinguishing}, ER-ZSAR \cite{chen2021elaborative} and VGGishZSL \cite{xie2021zero} are three typical semantic-based baselines, which are specifically designed for HAR, audio recognition, and image recognition.
They all connect the sensor data's embedding with the semantic embeddings of classes or sentence descriptions generated from language models such as Word2Vec \cite{mikolov2013distributed} and BERT \cite{devlin2018bert}.

\textit{GAN-based approaches.} 
TF-VAEGAN \cite{narayan2020latent} is a GAN-based open-set recognition approach, which uses a semantic decoder to synthesize features for unseen classes.
As current GAN-based approaches mainly focus on vision tasks, they are not used for the comparison on audio-related tasks.




\section{Evaluation}

\subsection{Applications and Datasets}
We evaluate EdgeFM on three application scenarios, i.e., HAR, robotic semantic sensing, and audio recognition.
Table~\ref{tab:datasets} shows the details of the five datasets and their corresponding FM.

\subsubsection{Human Activity Recognition}
\hfill\\
\noindent\textbf{Self-collected HAR dataset (SC15).}
We collect a real-world HAR dataset in an indoor home setting as shown in Figure~\ref{fig:testbed_AD}.
The dataset contains 30 subjects and each subject performs 15 activities, including sleeping, sitting on a chair, standing, squatting, playing mobile phone, rummaging cabinet, eating, playing chess, drinking, playing computer, reading books, watering flowers, handwashing, sweeping the floor, and brushing teeth.
The total duration of the dataset is about 20 hours, where the sampling rate is 20Hz. 
The collected video data is split into 2-second recordings, and the middle frame is selected as input samples of EdgeFM, which is the same setting in \cite{radford2021learning, girdhar2023imagebind}.

\noindent
\textbf{UCF101}\cite{soomro2012ucf101}.
This is a public video-based human activity recognition dataset.
We extract the middle frame of each activity video as input samples, which is the same processing way as in \cite{radford2021learning, girdhar2023imagebind}. 
It contains 11,331 images from 101 different categories of human activities.
We use the same train-test split as in \cite{radford2021learning} for a fair evaluation.

\subsubsection{Robotic Semantic Sensing}
\begin{figure}
    \centering

    \begin{subfigure}{1\columnwidth}
        \centering
        \includegraphics[width=\textwidth]{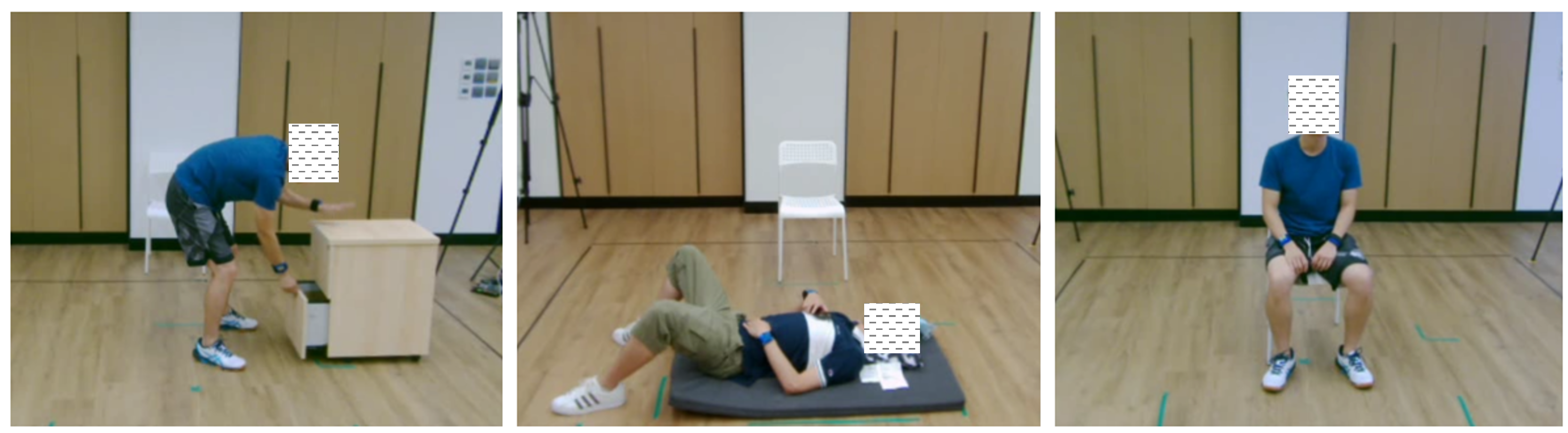}
        \caption{HAR testbed setup.}  \label{fig:testbed_AD}
    \end{subfigure}
    \begin{subfigure}{1\columnwidth}   
        \centering 
        \includegraphics[width=\textwidth]{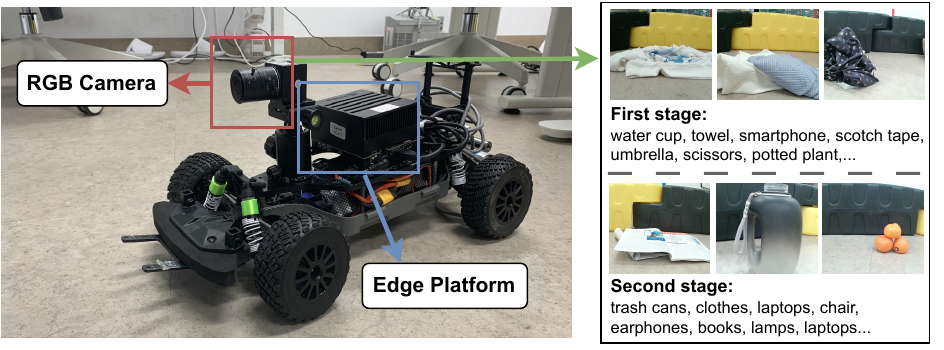}
        \caption{Robot semantic sensing testbed setup.}    \label{fig:testbed_robot}
    \end{subfigure}
     \vspace{-2em}
  \caption{Two real-world testbed setups.}
  \label{fig:testbed}
    \vspace{-2em}

\end{figure}

\hfill\\
\noindent
\textbf{Self-collected indoor scene dataset (SC40)}.
We collect a real-world indoor scene dataset,
where objects in the environment are allowed to be added dynamically.
As shown in Figure~\ref{fig:testbed_robot}, a robot equipped with a camera and edge platform moves randomly in the environment and takes RGB images continuously.
We collect 18,295 RGB images in 40 classes in total.
During the data collection, we place half of the classes first and place the other later.
Current indoor scene datasets like SUN RGB-D \cite{song2015sun} do not consider the scenario where users dynamically add objects into the environment, which is the main purpose of our self-collected dataset.


\noindent
\textbf{FLO102} \cite{nilsback2008automated}.
This is a public dataset that contains 8,189 images from 102 different categories of flowers.
We use the same train-test split as in \cite{radford2021learning} for a fair evaluation.

\subsubsection{Audio Recognition}
\hfill\\
\noindent
\textbf{ESC50} \cite{piczak2015esc}.
This is a public audio dataset that contains 2,000 audio clips from 50 different classes of environmental sound.
Each audio clip is a 5-sec recording sampled at 44.1 kHz. 
We use the same train-test split as in \cite{xie2021zero} for a fair evaluation.




\begin{table}
  \caption{Details of datasets, and FMs used in evaluation.
  CLS is the number of classes.}
  \label{tab:datasets}
  \begin{tabular}{cccl}
    \toprule
    Dataset & Tasks & CLS & FMs \\
    \midrule
    SC15& Activity recognition     & 15 & ImageBind/CLIP \\    
    UCF101& Activity recognition     & 101 & ImageBind/CLIP \\
    SC40& Indoor scene recognition     & 40 & ImageBind/CLIP\\
    FLO102& Flower recognition & 102 & ImageBind/CLIP\\
    ESC50& Audio recognition  & 50  & ImageBind\\
  \bottomrule
\end{tabular}
\end{table}


\subsection{An End-to-End Application}
We conduct an end-to-end test by deploying EdgeFM with CLIP as the FM on a mobility robot for semantic sensing. 
Figure~\ref{fig:dynamic_threshold_map} shows the moving trajectory of the robot in a room.

\subsubsection{Adaptability to Network Variation}
\label{Adaptability to Network Variation}
We evaluate the adaptability of EdgeFM to network variation.
When the robot moves, the network bandwidth fluctuates with time and location, where the lowest and the highest bandwidth are 2 Mbps and 123 Mbps, respectively.
We prioritize the execution performance by setting the latency bound to 30ms, which can meet real-time requirements for most of the applications \cite{huang2022real}.
EdgeFM tunes the threshold of model switching according to the bandwidth in real-time, ranging from 0 to 1 with an interval of 0.05.

\begin{figure}
    \centering
    \begin{subfigure}{0.88\columnwidth}
        \centering
        \includegraphics[width=\textwidth]{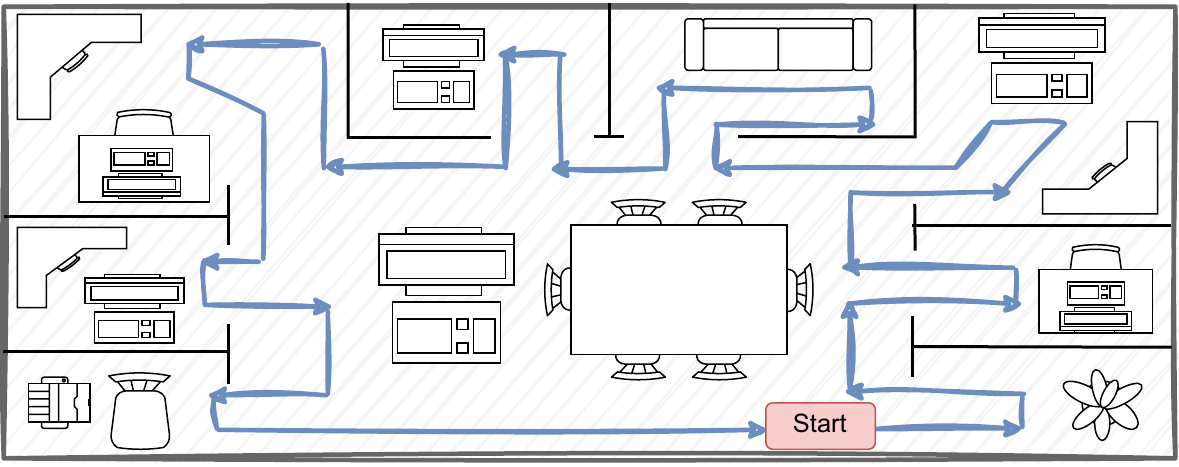}
        \caption{Moving trajectories of the robot in the room.}  \label{fig:dynamic_threshold_map}
    \end{subfigure}
     \vspace{1.0em}
    \begin{subfigure}{1\columnwidth}   
        \centering 
        \includegraphics[width=\textwidth]{{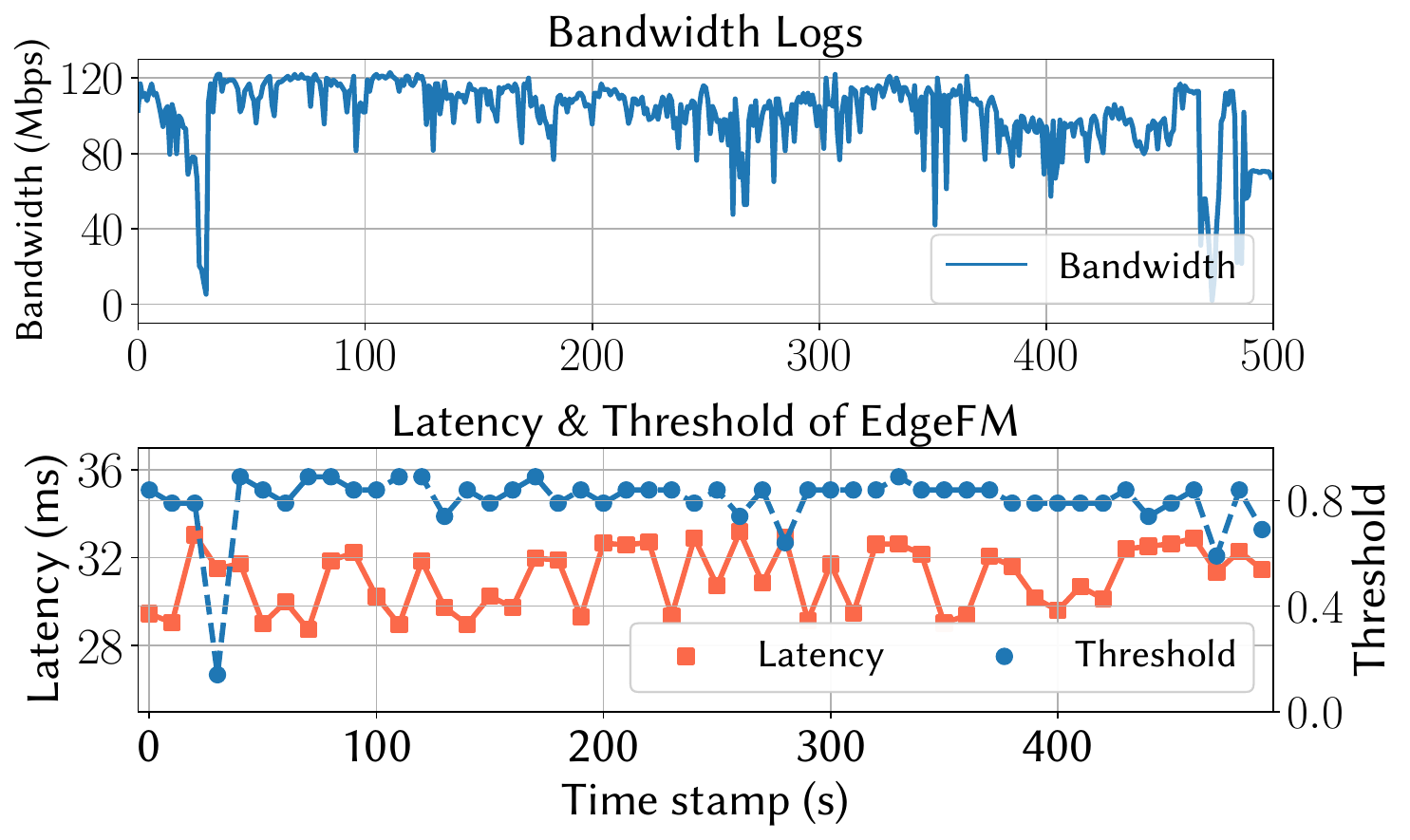}}
        \caption{Variation of network bandwidth and the threshold and inference latency of EdgeFM.}
        \label{fig:dynamic_threshold_bandwidth}
    \end{subfigure}
  \caption{EdgeFM's setup and system indicators of the end-to-end evaluation.}
  \label{fig:dynamic_threshold}
\end{figure}

Figure~\ref{fig:dynamic_threshold_bandwidth} shows that EdgeFM sets the threshold to a relatively high value ($\sim$0.99) to ensure that most of the samples are offloaded to the cloud for inference, at high bandwidth conditions (e.g., $t\in[50,200]$).
When the bandwith is low (e.g., $t\in[20,50]$), EdgeFM sets the threshold to a relatively low value ($\sim$0.15) to make most data processed on the edge while only few samples are offloaded to the cloud.
Overall, results show that EdgeFM can successfully adapt to the dynamic network variation by tuning the threshold of model switching at runtime.

\subsubsection{Adaptability to Environment Change}
\label{E2E_domain_adaptation}
We further evaluate the adaptability of EdgeFM to the environment change, i.e. both data distribution and interested classes change.
To simulate the scenario where items in users' homes change over time, we first add half of the classes into the environment and then add the remaining classes later.
We run EdgeFM continuously in an unsupervised manner to evaluate the adaptability to the environment change.

\begin{figure}[h]
  \centering
\includegraphics[width=1\linewidth]{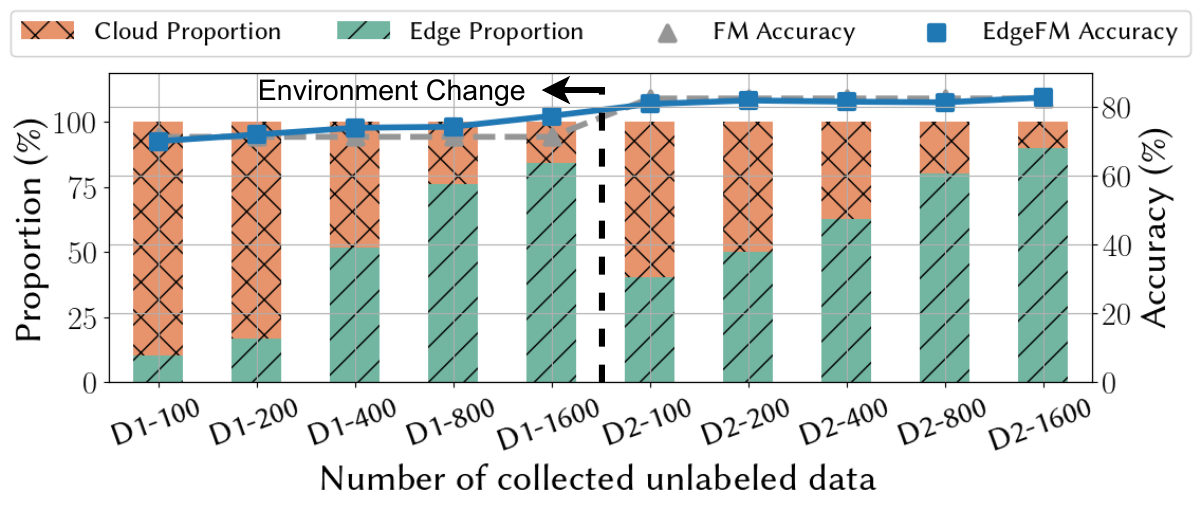}
  \caption{EdgeFM's performance when the environment changes in the end-to-end evaluation. X-axis shows the amount of collected data. D1 and D2 denote the two environments.}
  \label{fig:domain_change_FLO}
\end{figure}

Figure~\ref{fig:domain_change_FLO} shows the proportion of edge-cloud processed data, the overall accuracy of EdgeFM and the original FM, and the moment when environment change occurs.
The grey dashed line increases after the environment change as the original FM has higher accuracy for the second half of classes.
The result shows that EdgeFM can adjust the proportion of edge-cloud processed data at runtime to adapt to the environment change.
To maintain the overall accuracy close to the original FM's accuracy, the edge processing proportion of EdgeFM decreases from 84.4\% to 40.2\% after environment change (i.e., the green bar in Figure~\ref{fig:domain_change_FLO}).
Compared with traditional close-set approaches, EdgeFM can reduce the efforts of manual labeling significantly.

\subsection{Overall Performance of EdgeFM}

\subsubsection{Comparison with Baselines of Efficient On-device Inference}

We evaluate both the inference latency and accuracy of EdgeFM and on-device NN efficient inference baselines.
We implement PersEPhonEE \cite{leontiadis2021s}
and SPINN \cite{laskaridis2020spinn} on two FMs, i.e., ImageBind and CLIP.
We set the bandwidth to 55 Mbps for all tests in this evaluation. 

\begin{figure}
    \centering
    \begin{subfigure}{0.48\columnwidth}
        \centering
        \includegraphics[width=\textwidth]{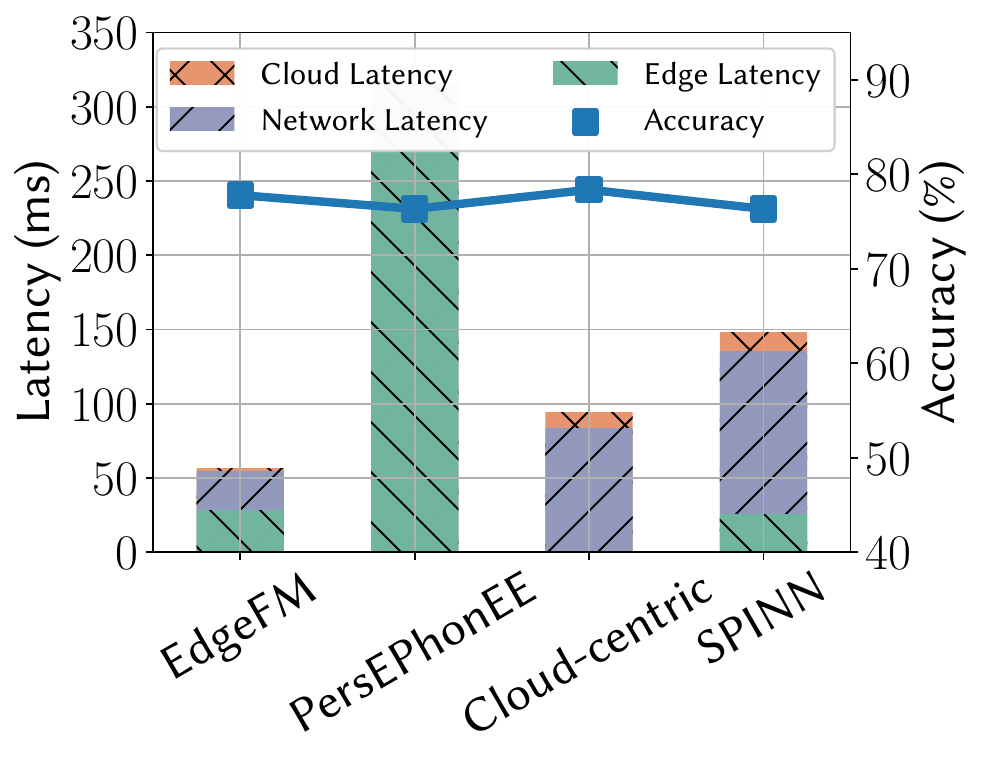}
        \caption{ImageBind, FLO102.}  \label{fig:imagebind_flo}
    \end{subfigure}
    \hfill
    \begin{subfigure}{0.48\columnwidth}  
        \centering 
        \includegraphics[width=\textwidth]{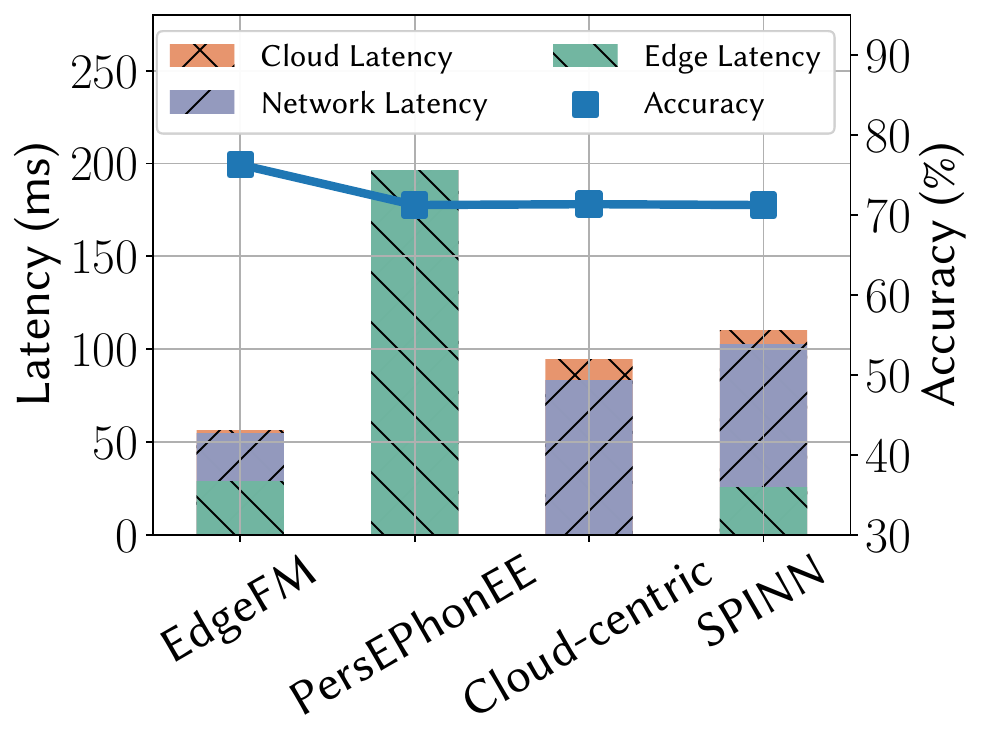}
        \caption{ImageBind, SC40.}    
        \label{fig:imagebind_cpii}
    \end{subfigure}
    
    \vskip\baselineskip
    \begin{subfigure}{0.48\columnwidth}   
        \centering 
        \includegraphics[width=\textwidth]{{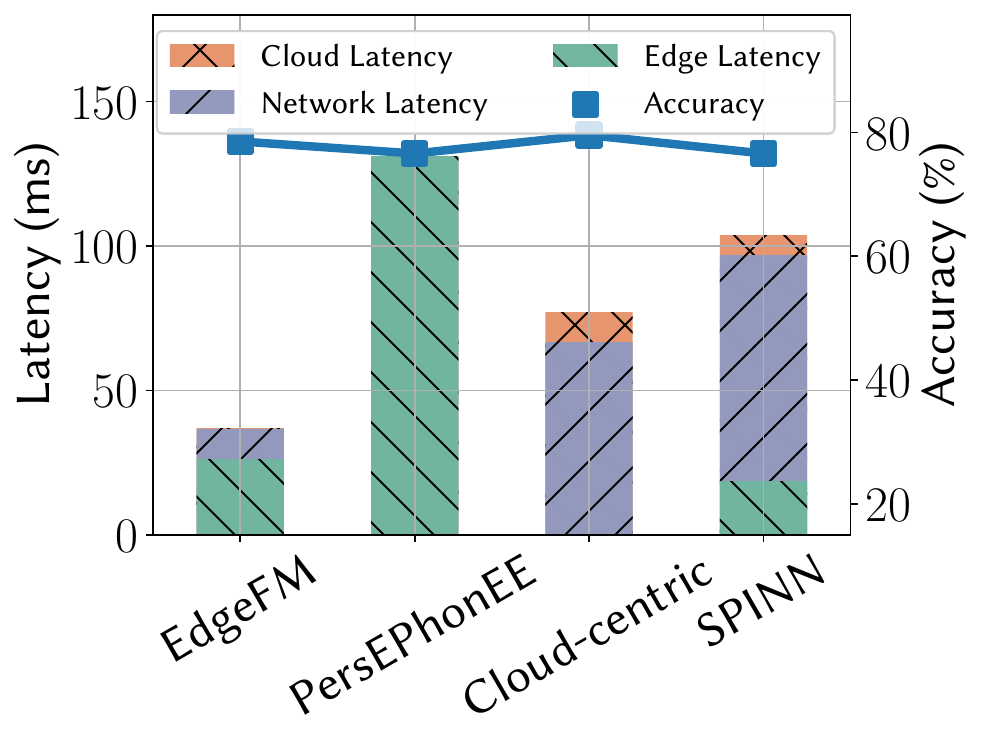}}
        \caption{CLIP, FLO102.}    \label{fig:clip_flo}
    \end{subfigure}
    \hfill
    \begin{subfigure}{0.48\columnwidth}   
        \centering 
        \includegraphics[width=\textwidth]{{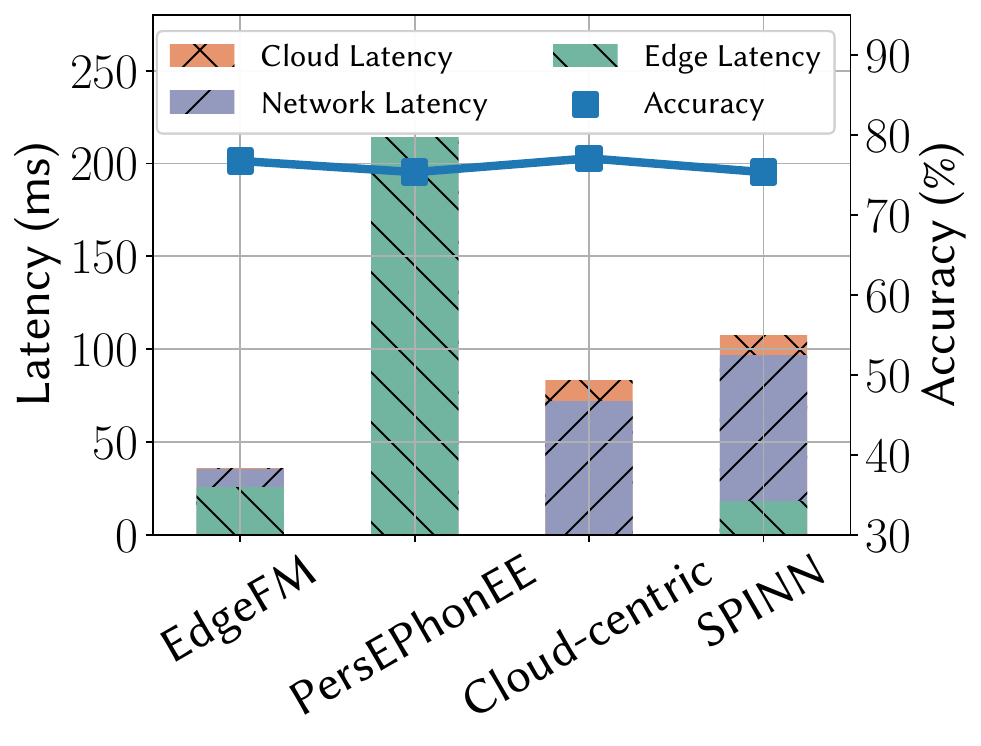}}
        \caption{CLIP, SC40.}    \label{fig:clip_cpii}
    \end{subfigure}

    \caption{Accuracy and inference latency of EdgeFM and other on-device NN efficient inference systems.}
    \label{fig:overall_performance}

\end{figure}

The results in Figure~\ref{fig:overall_performance} and Table~\ref{tab:overallperformance} show that EdgeFM can achieve up to \textbf{1.52x}$\sim$\textbf{2.63x} end-to-end latency reduction compared with the best baseline approaches for ImagBind, and can also achieve up to \textbf{1.27x}$\sim$\textbf{3.22x} end-to-end latency reduction compared with the best baseline approaches for CLIP.
As shown in Figure~\ref{fig:overall_performance}, among these approaches, only the latency of EdgeFM is lower than 60ms under the 55 Mbps bandwidth, which can meet the real-time requirements for most applications \cite{meng2023enabling}.
Meanwhile, EdgeFM can achieve higher accuracy than both SPINN and PersEPhonEE on the two FMs.
This is because PersEPhonEE adopts an early-exit mechanism to reduce redundant computation. 
However, early-exit heads on FMs are heavyweight due to the high-dimensional embedding of FMs and deep layers.
Although SPINN can offload computation to the cloud, it needs to transmit a large size intermediate embeddings. 
The size of intermediate embeddings of ImageBind is 257$\times$1$\times$1280, which is much larger than the size of the raw image (i.e., 3$\times$224$\times$224).

Another observation is that EdgeFM's recognition accuracy can even outperform the cloud-centric approach on some certain datasets (see Figure~\ref{fig:imagebind_cpii}).
This is because a dedicated SM trained with abundant data (e.g., more than 800) can outperform FMs, which aligns with the findings shown in Figure~\ref{fig:motivation}.
However, for more challenging datasets with more diverse data, such as UCF101, the performance of customized small models remains inferior to that of the FMs.
This discrepancy also proves the remarkable generalization ability of FMs.


\begin{table}[t]
  \caption{The reduced end-to-end latency compared with the best baselines on two edge platforms. 
  N.A. means not applicable since CLIP supports vision data only.}
  \label{tab:overallperformance}
  \begin{tabular}{ccccccl}
    \toprule
   FM &Device& FLO102 & UCF101 & SC40 & SC15&ESC50 \\
    \midrule
   
   \multirow{2}{*}{ImageBind}&Xavier &1.67x& 2.63x& 1.65x  & 1.52x &1.70x\\   
   &Nano &1.80x& 2.36x& 1.84x & 1.73x &1.32x\\  
   \midrule

   \multirow{2}{*}{CLIP}&Xavier &2.10x& 2.20x& 2.36x & 2.01x &N.A.\\   
   &Nano &2.52x& 1.27x& 3.22x & 2.37x &N.A.\\ 
   
  \bottomrule
\end{tabular}
\end{table}

   

   

   



\subsubsection{Impact of Network Bandwidth}
\begin{figure}
    \centering
    \begin{subfigure}{0.48\columnwidth}
        \centering
        \includegraphics[width=\textwidth]{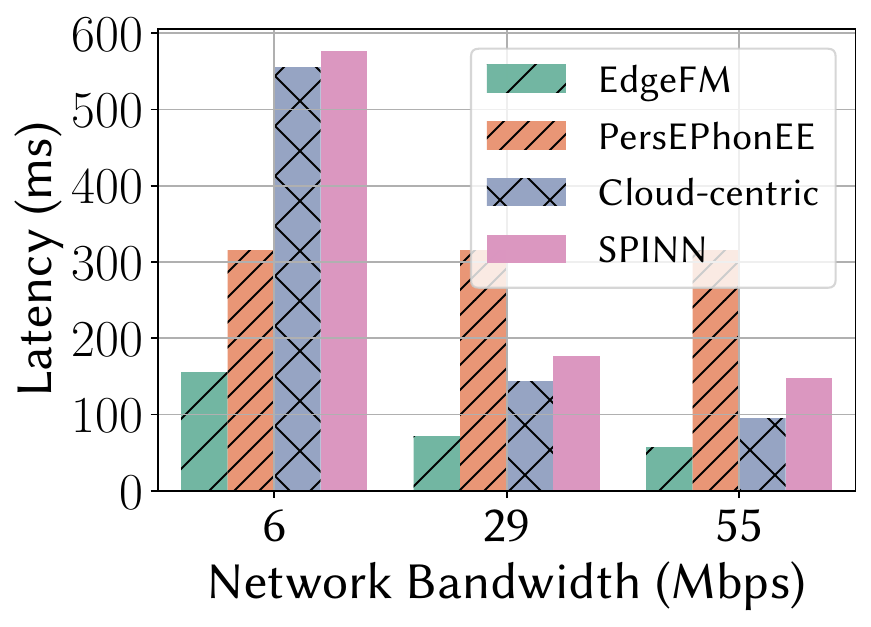}
        \caption{ImageBind, FLO102.}  \label{fig:bandwidth_imagebind_flo}
    \end{subfigure}
    \hfill
    \begin{subfigure}{0.48\columnwidth}  
        \centering 
        \includegraphics[width=\textwidth]{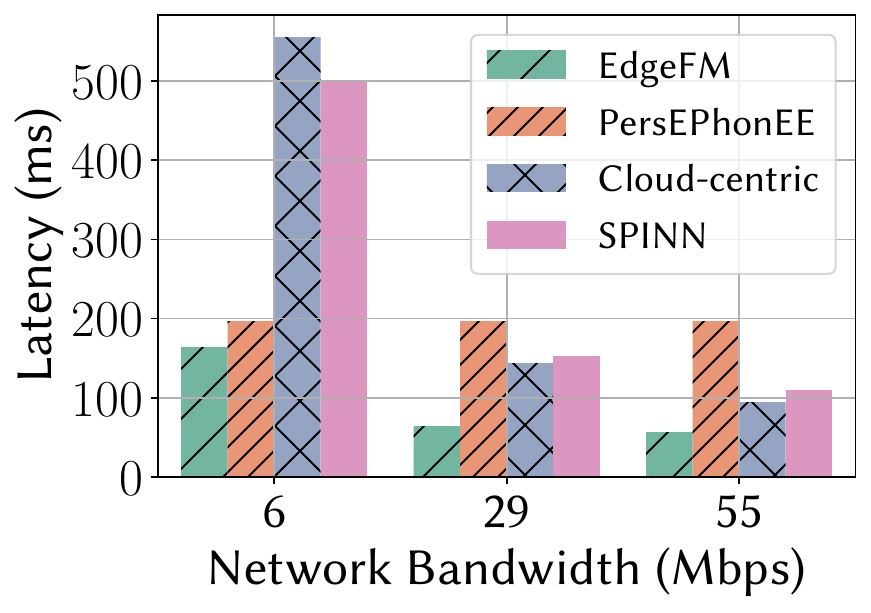}
        \caption{ImageBind, SC40.}    
        \label{fig:bandwidth_imagebind_cpii}
    \end{subfigure}
    \vskip\baselineskip
    \begin{subfigure}{0.48\columnwidth}   
        \centering 
        \includegraphics[width=\textwidth]{{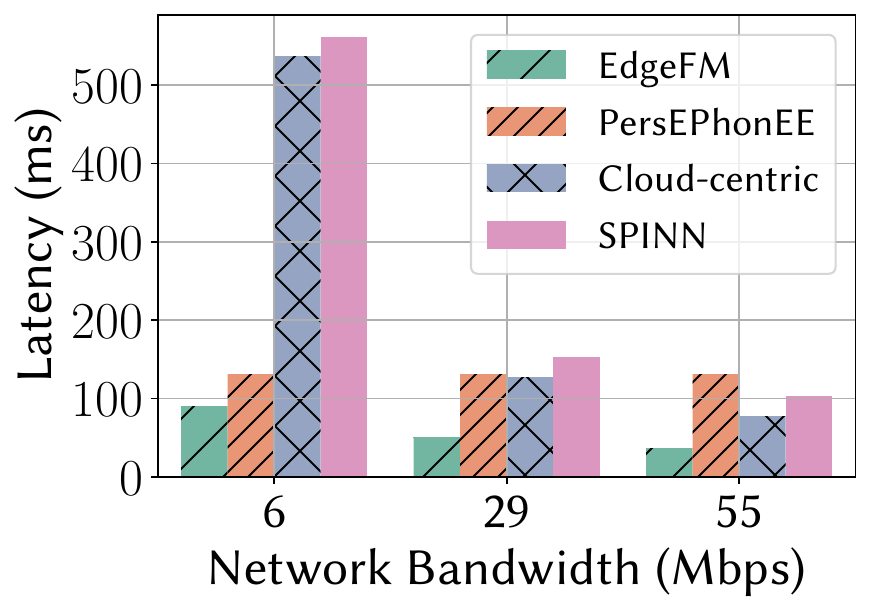}}
        \caption{CLIP, FLO102.}    \label{fig:bandwidth_clip_flo}
    \end{subfigure}
    \hfill
    \begin{subfigure}{0.48\columnwidth}   
        \centering 
        \includegraphics[width=\textwidth]{{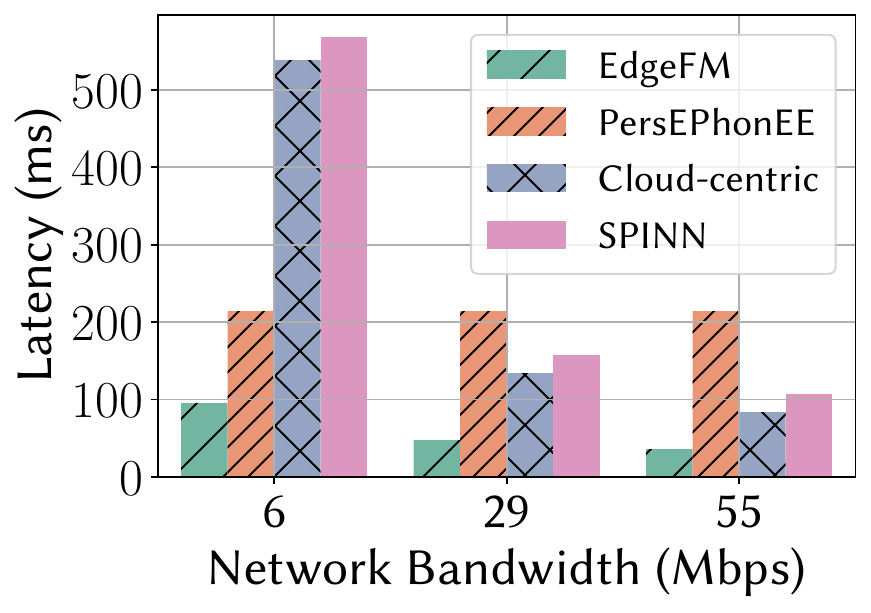}}
        \caption{CLIP, SC40.}    \label{fig:bandwidth_clip_cpii}
    \end{subfigure}
    \caption{Impact of network bandwidth on EdgeFM.}
    \label{fig:network_bandwidth}
\end{figure}

We evaluate the impact of network bandwidth on EdgeFM.
We conduct evaluations under low (6 Mbps), middle (29 Mbps), and high (55 Mbps) network bandwidth, respectively.
Figure~\ref{fig:network_bandwidth} shows that EdgeFM can achieve up to \textbf{3.5x} and \textbf{3.7x} end-to-end inference speedup on ImageBind compared with cloud-centric and SPINN under low network conditions (6 Mbps).
Under high network bandwidth conditions, this gap narrows, where EdgeFM is still able to achieve up to \textbf{1.7x} and \textbf{2.4x} end-to-end inference speedup compared with cloud-centric and SPINN under 55 Mbps bandwidth. 
Results in Figure~\ref{fig:network_bandwidth} show that EdgeFM performs better than the existing solutions, especially under low bandwidth conditions.

\subsubsection{Comparison with Open-set Recognition Approaches}
We then evaluate the open-set recognition accuracy and inference latency of EdgeFM and open-set recognition approaches on two edge platforms.
The network bandwidth is also set to 55 Mbps for EdgeFM.

As shown in Table~\ref{tab:ZSL_CLIP}, EdgeFM achieves \textbf{26.7\%} higher accuracy on average than GAN-based approaches.
Compared with semantic-based approaches, EdgeFM achieves \textbf{21.2\%} and \textbf{21.7\%} higher accuracy than DUS-VAE and ES-ZSAR, respectively.
Since EdgeFM adopts lightweight small models on the edge, the end-to-end inference latency of EdgeFM is lower than the baselines. 
For Jetson Nano, EdgeFM reduces the latency by 1.73x and 1.98x compared with the best baseline on FLO102 and UCF101 datasets respectively.

We also implement EdgeFM on the audio branch of ImageBind and compare it with baselines on the audio recognition task.
EdgeFM achieves \textbf{34.3\%} accuracy gain and 3.22x inference latency reduction on Jetson Nano compared with VGGishZSL on ESC dataset. 
VGGishZSL adopts VGG19 \cite{simonyan2014very} as the audio feature extractor, where its FLOPS and parameters are 10x larger than the lightweight, small model used in EdgeFM.
However, the inference latency of EdgeFM is slower than VGGishZSL 12.9ms on Xavier.
This is because the strong computing power of Jetson AGX Xavier's GPU makes the parallel computation highly efficient, 
which causes the latency to be not equivalent to the FLOPS and parameters.


\begin{table}
  \caption{Overall accuracy(\%) and latency of EdgeFM compared with other open-set recognition approaches.}
  \label{tab:ZSL_CLIP}
  \begin{tabular}{ccccl}
    \toprule
 Dataset &   Approach  & Acc & Xavier &  Nano\\
    \midrule

\multirow{3}{*}{FLO102} & TF-VAEGAN \cite{narayan2020latent}&  62.5\% & 54.2 ms   & 108.1 ms \\

& DUS-VAE \cite{su2022distinguishing}&  62.1\% & 57.7 ms   & 110.1 ms \\

& \textbf{EdgeFM} &  \textbf{83.3\%} & \textbf{44.6 ms} & \textbf{62.2 ms} \\

\midrule

\multirow{3}{*}{UCF101} & TF-VAEGAN \cite{narayan2020latent} &  41.0\% & 53.9 ms   & 107.1 ms\\

& ER-ZSAR \cite{chen2021elaborative} &  51.8\% & 87.9 ms   & 424.8 ms \\

& \textbf{EdgeFM} &  \textbf{73.5\%} & \textbf{42.7 ms}   & \textbf{54.0 ms} \\

\midrule
\multirow{2}{*}{ESC50} & VGGishZSL \cite{xie2021zero} & 33.0\% & 42.2 ms   & 217.2 ms \\

& \textbf{EdgeFM}&  \textbf{67.3\%} & \textbf{ 55.1 ms}   & \textbf{67.5ms} \\

  \bottomrule
\end{tabular}
      \vspace{1em}
\end{table}

\subsection{Understanding EdgeFM's Performance}

\subsubsection{Proportion of the Data Processed on Edge}
\begin{figure}
    \centering
    \begin{subfigure}{0.495\columnwidth}
        \centering
\includegraphics[width=\textwidth]{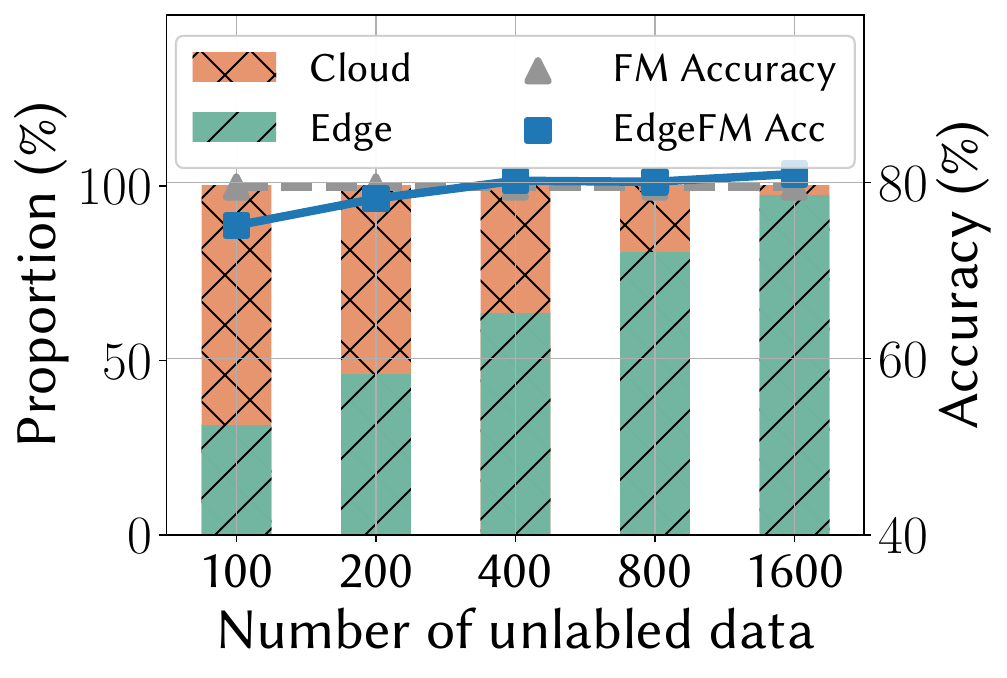}
         \caption{FLO102.}  \label{fig:data_ratio_flo}
    \end{subfigure}
    \hfill
    \begin{subfigure}{0.495\columnwidth}  
        \centering 
\includegraphics[width=\textwidth]{{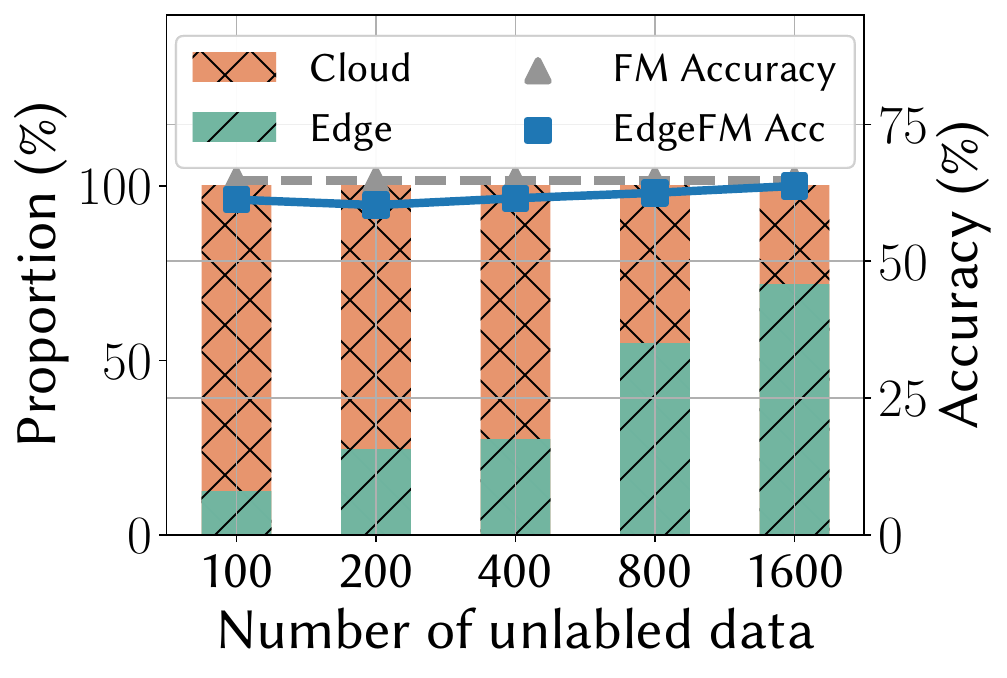}}
        \caption{ESC50.}    \label{fig:data_ratio_esc}
    \end{subfigure}
  \caption{Proportion of the data processed by the customized small model on edge and FM on the cloud.}
  \label{fig:data_ratio}
\end{figure}

We assess the proportion of the data processed by the customized small model on edge and the FM on cloud in EdgeFM, where we use CLIP as the FM in the experiments.
Figure~\ref{fig:data_ratio} shows that the proportion of the edge processed data increases when more data is collected.
The proportion of data processed on edge increases from 31.1\% to 63.5\% when the collected data increases from 100 to 400.
The proportion can increase up to 97.3\% when collecting 1600 samples in the environment, which means only 2.7\% data are required to be uploaded to the cloud for inference, thus can reduce end-to-end latency compared with cloud-centric solutions.
Moreover, our dynamic model switching strategy can keep the overall accuracy always close to the original FM.



\subsubsection{Effectiveness of Semantic-driven Customization.}
\begin{figure}
    \centering
    \begin{subfigure}{0.49\columnwidth}
    
        \centering
        \includegraphics[width=\textwidth]{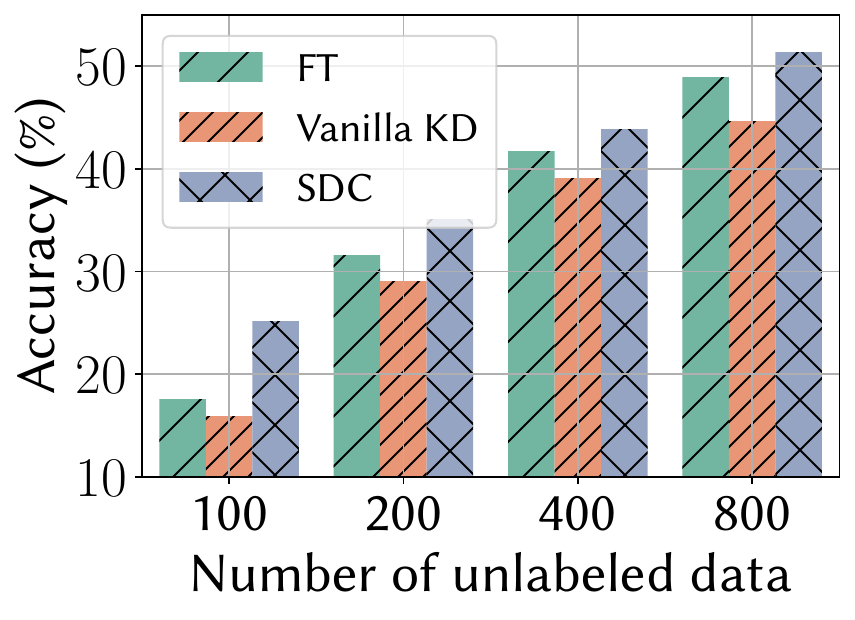}
        \caption{Edge-only performance.}  \label{edgemodel}
    \end{subfigure}
    \hfill
    \begin{subfigure}{0.49\columnwidth}
        \centering 
        \includegraphics[width=\textwidth]{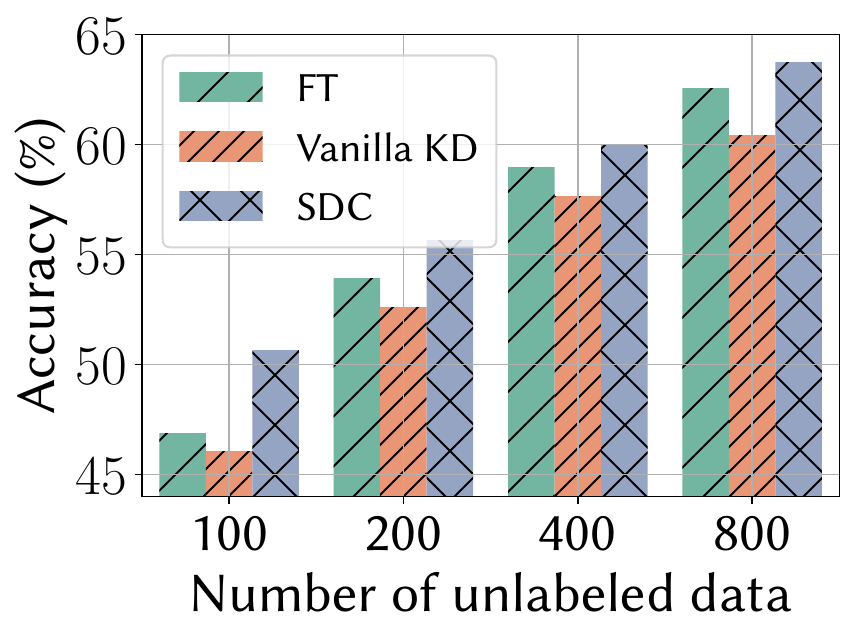}
        \caption{Edge-cloud performance.}  \label{edgecloud}
    \end{subfigure}
  \caption{The performance of EdgeFM's semantic-driven customization.}
  \label{fig:effectiveness_kd}
\end{figure}

We compare our semantic-driven customization with the vanilla KD and fine-tuning with the hard pseudo label (FT).
The vanilla KD \cite{hinton2015distilling} adopts the standard KL divergence to minimize the embedding gap between FMs and small models without using the pseudo text embeddings from FMs.
FT refers to adopting the hard pseudo label predicted by FMs as ground truth and cross-entropy for distillation.
Figure~\ref{edgemodel} shows that our semantic-driven customization (marked as SDC in the figure) is able to achieve up to 9.2\%, 6.1\%, 4.7\%, and 6.7\% accuracy gain for small model performance under the different number of training data.
Since hard pseudo labels fail to preserve semantic relationships between categories \cite{islam2019soundsemantics,tong2021zero}, FT performs inferior to our approach.
Vanilla KD fails to leverage the knowledge within text embeddings, thus resulting in inferior performance compared to our approach.
Figure~\ref{edgecloud} shows the edge-cloud performance between the three approaches, where we keep the data uploading proportion the same (50\% in our setting) for a fair comparison.
The result show that EdgeFM's semantic-driven customization can achieve up to 4.6\%, 3.0\%, 3.0\%, 2.4\%, and 3.4\% accuracy gain under the different number of training data.

\subsubsection{Trade-off between Accuracy and Latency}
\begin{figure}
    \centering
    \begin{subfigure}{0.495\columnwidth}
        \centering
        \includegraphics[width=\textwidth]{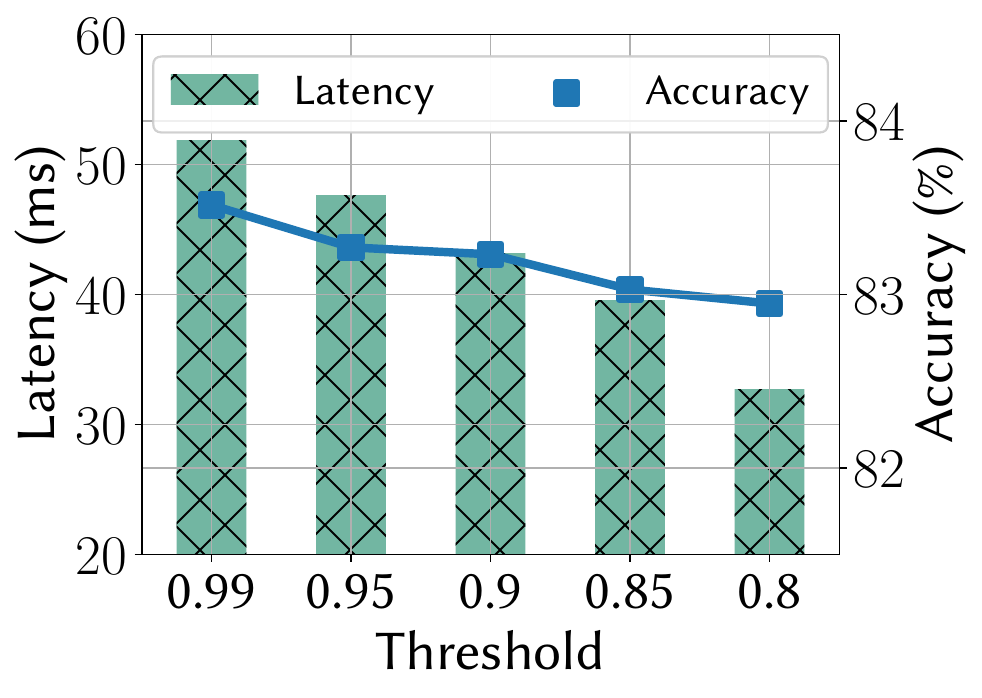}
        \caption{FLO102.}  
        \label{fig:trade-off_flo}
    \end{subfigure}
    \hfill
    \begin{subfigure}{0.495\columnwidth}  
        \centering 
        \includegraphics[width=\textwidth]{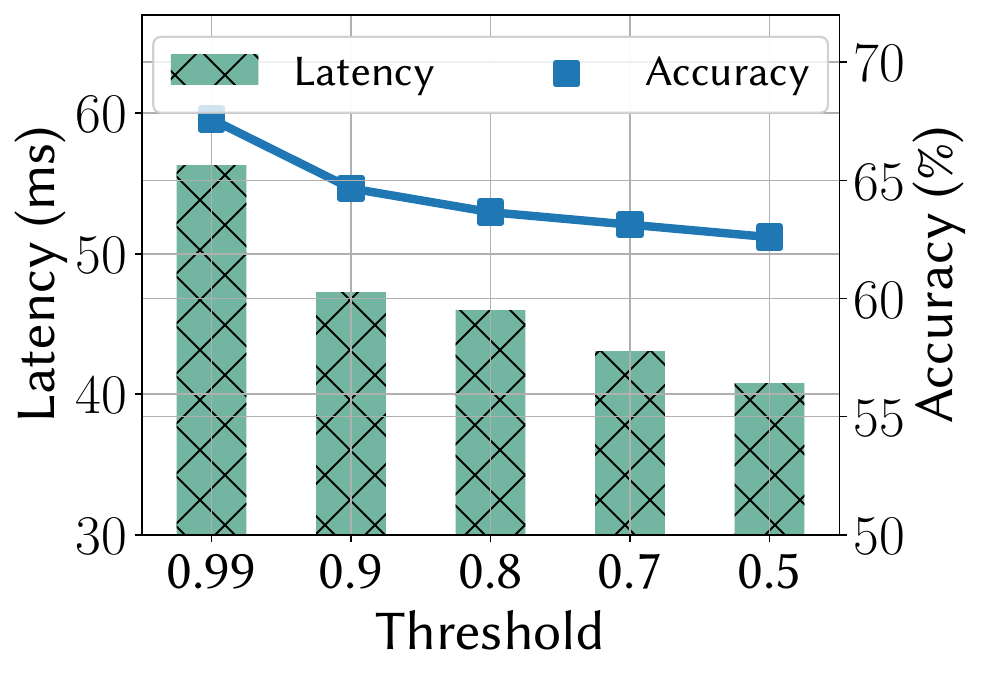}
        \caption{UCF101.}    
        \label{fig:trade-off_ucf}
    \end{subfigure}
  \caption{Accuracy-latency trade-off caused by the confidence threshold in EdgeFM.}
  \label{fig:trade-off}
\end{figure}

Figure~\ref{fig:trade-off} shows the accuracy-latency trade-off caused by the threshold of model switching in EdgeFM.
Setting a higher threshold leads to more sensor data offloaded to the cloud, and higher overall accuracy but longer end-to-end inference latency.
On the other hand, a lower threshold makes 
more sensor data processed by the customized small models on the edge and save the network transmission latency, achieving faster end-to-end inference but lower accuracy.
Therefore, there is a trade-off between accuracy and latency caused by the model switching thresholds.
EdgeFM adjusts the threshold at runtime considering the variation of network bandwidth.
The evaluation results can be found in Section~\ref{Adaptability to Network Variation}.

\section{Discussion}
\textbf{Scalability to Other FMs.}
This work targets multi-modal FMs based on embedding matching paradigms. EdgeFM deploys CLIP and ImageBind, which are the most popular FMs in the category.
There are recent works proposed based on the similar matching method, such as DetCLIPv2 \cite{yao2023detclipv2} for detection, SAM \cite{kirillov2023segment} for segmentation.
We envision that EdgeFM's collaboration between the FMs and specialized small models and iteratively querying specific knowledge from FMs can be extended to other FMs.

\noindent\textbf{Change of Distribution.}
The machine learning models can be formulated as $\mathcal{Y}=f(\mathcal{X})$. 
There is a set of existing works \cite{akbari2019transferring,luo2021phyaug,gong2019metasense} study the change of data distribution $\mathcal{X}$.
This paper focuses on open-set learning on the edge, i.e., the dynamic change of class set $\mathcal{Y}$.
In fact, the change of interested class set $\mathcal{Y}$ can be regarded as a generalized distribution change, including the change of both $\mathcal{X}$ and $\mathcal{Y}$.
Our evaluation in Section~\ref{E2E_domain_adaptation} implicitly shows that EdgeFM can also well adapt to the distribution changes.

\noindent\textbf{The Optimal Choice of Small Models.}
EdgeFM pre-stores a model pool containing many common small model architectures with diverse accuracy, FLOPS, memory, and latency.
This pool also includes architectures through Neural Architecture Search (NAS) \cite{elsken2019neural}, such as EfficientNet \cite{tan2019efficientnet}.
Recent studies \cite{9156803} have investigated the utilization of NAS to look for the best student architecture for the given teacher model during the KD process. 
They can be integrated with EdgeFM to search for the most suitable small model architecture for a given FM.


\noindent\textbf{Applications with Labeled Calibration Data.}
EdgeFM focuses on the applications without labeled data for calibration to eliminate the overhead of manual labelling.
Recent studies have shown that the knowledge in FMs can be better evoked via parameter-efficient fine-tuning (PEFT) approaches \cite{hu2021lora}. 
EdgeFM also supports working in the scenario when labeled data is available.
In such a case, EdgeFM first uses the labeled data to fine-tune FMs by PEFT on the cloud.
Then, the fine-tuned FM can further provide a knowledge base service for small models to query.

\noindent\textbf{Scalability to Other Sensor Modalities.}
EdgeFM supports other time-series sensor data such as video, audio, and IMU.
FMs used by EdgeFM, i.e., ImageBind, can support diverse sensor modalities with the corresponding pre-trained encoders.
The techniques for video streaming such as frame filtering \cite{chen2015glimpse}, can also be integrated with EdgeFM to improve the efficiency of processing time-series vision data.

\section{Conclusion}


This paper proposes EdgeFM, a novel edge-cloud cooperative system which empowers embedded systems with open-set recognition ability by leveraging FMs for selective knowledge query and edge model customization.
EdgeFM maintains the overall performance always close to the FM by dynamic model switching.
Extensive experiments show that EdgeFM can reduce the end-to-end latency up to 3.2x and achieve 34.3\% accuracy increase compared with the baseline.


\section{Acknowledgement}
This paper is supported in part by the Research Grants Council (RGC) of Hong Kong under Collaborative Research Fund (CRF) grants C4072-21G and C4034-21G, General Research Fund (GRF) 14214022, Faculty of Engineering of The Chinese University of Hong Kong under Direct Grant 4055167, National Science Foundation of China (NSFC) under Young Scientists Fund 62202407.


\balance
\bibliographystyle{ACM-Reference-Format}
\bibliography{EdgeFM}
\balance










\end{document}